\documentclass[10pt,twocolumn,letterpaper]{article}

\usepackage{cvpr}              
\usepackage{multirow}
\usepackage{booktabs}
\usepackage[table]{xcolor}
\usepackage[normalem]{ulem} 
\usepackage{tcolorbox}
\newcommand{\tboxsize}{
  \fontsize{8pt}{10pt}\selectfont
}
\definecolor{cvprblue}{rgb}{0.21,0.49,0.74}
\definecolor{lightgray}{gray}{0.9}
\usepackage[pagebackref,breaklinks,colorlinks,allcolors=cvprblue]{hyperref}

\def\paperID{13882} 
\def\confName{CVPR}
\def\confYear{2026}
\newcommand{\method}{Reward Forcing}
\title{Reward Forcing: Efficient Streaming Video Generation with \\Rewarded Distribution Matching Distillation}

\author{Yunhong Lu$^{1,2}$ {\quad} Yanhong Zeng$^{\dagger, 2}$ {\quad} Haobo Li$^{2,4}$ {\quad} Hao Ouyang$^2$ {\quad} Qiuyu Wang$^2${\quad} \\ Ka Leong Cheng$^2$  {\quad}
Jiapeng Zhu$^2$ {\quad} Hengyuan Cao$^1$ {\quad} Zhipeng Zhang$^4$ {\quad} \\ Xing Zhu$^2$ {\quad} Yujun Shen$^2$ {\quad} Min Zhang$^{*,1,3}$ 
\\
$^1$Zhejiang University {\quad} $^2$Ant Group {\quad} $^3$SIAS-ZJU {\quad} $^4$SJTU\\
}

\newcounter{suppsection}
\renewcommand{\thesuppsection}{S\arabic{suppsection}}

\newcommand{\suppsection}[1]{%
    \refstepcounter{suppsection}%
    \section*{\thesuppsection: #1}%
    \addcontentsline{toc}{section}{Supplementary Section \thesuppsection: #1}%
    \label{suppsec:\thesuppsection}
}

\begin{document}
\twocolumn[{
\renewcommand\twocolumn[1][]{#1}
\maketitle

\begin{center}
    \vspace{-0.7cm} 
    \ttfamily \href{https://reward-forcing.github.io/}{https://reward-forcing.github.io/}
    \vspace{0.2cm}  
\end{center}

\vspace{-0.8cm}
\begin{center}
    \captionsetup{type=figure}
    \includegraphics[width=1\linewidth]{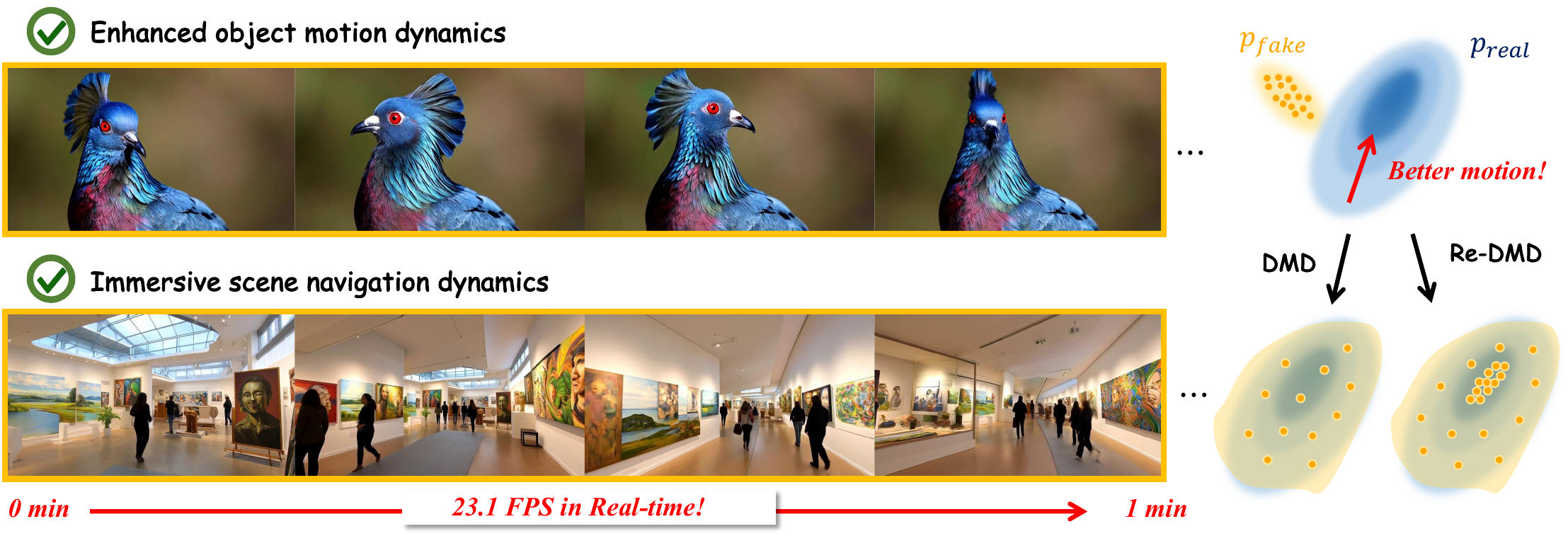}
    \captionof{figure}{We propose \textbf{{\method}} to distill a bidirectional video diffusion model into a few-step autoregressive student model that enables real-time (23.1 FPS) streaming video generation. Instead of using vanilla distribution matching distillation (DMD), {\method} adopts a novel rewarded distribution matching distillation (Re-DMD) that prioritizes matching towards high-reward regions, leading to enhanced object motion dynamics and immersive scene navigation dynamics in generated videos.}
\label{fig:teaser}
\end{center}
}]

\let\thefootnote\relax\footnote{\scriptsize{* Corresponding author, $\dagger$ Project leader}}

\begin{abstract}
Efficient streaming video generation is critical for simulating interactive and dynamic worlds. Existing methods distill few-step video diffusion models with sliding window attention, using initial frames as sink tokens to maintain attention performance and reduce error accumulation. However, video frames become overly dependent on these static tokens, resulting in copied initial frames and diminished motion dynamics. To address this, we introduce \textbf{{\method}}, a novel framework with two key designs. First, we propose \textbf{EMA-Sink}, which maintains fixed-size tokens initialized from initial frames and continuously updated by fusing evicted tokens via exponential moving average as they exit the sliding window. Without additional computation cost, EMA-Sink tokens capture both long-term context and recent dynamics, preventing initial frame copying while maintaining long-horizon consistency. Second, to better distill motion dynamics from teacher models, we propose a novel \textbf{Rewarded Distribution Matching Distillation (Re-DMD)}. Vanilla distribution matching treats every training sample equally, limiting the model's ability to prioritize dynamic content. Instead, Re-DMD biases the model's output distribution toward high-reward regions by prioritizing samples with greater dynamics rated by a vision-language model. Re-DMD significantly enhances motion quality while preserving data fidelity. We include both quantitative and qualitative experiments to show that {\method} achieves state-of-the-art performance on standard benchmarks while enabling high-quality streaming video generation at \textbf{23.1 FPS} on a single H100 GPU.

\label{sec:abstract}
\end{abstract}

\section{Introduction}
\label{sec:intro}

The scaling of video diffusion transformers (DiTs) \cite{peebles2023scalable,wan2025wan} has advanced text-to-video generation, producing realistic videos with intricate dynamics \cite{videoworldsimulators2024,genie3,wiedemer2025video,cui2025self}. However, their simultaneous denoising of all frames using bidirectional attention hinders interactive applications, which demand streaming generation over extended horizons under strict latency constraints. Achieving both low latency and high visual-dynamic fidelity remains the central challenge.

To achieve efficient streaming video generation, recent advances distill slow pre-trained bidirectional diffusion models into efficient few-step autoregressive student models \cite{chen2025skyreels,yin2025slow,huang2025self}. In these models, each frame attends only to previous frames with sliding window attention, enabling real-time streaming inference through key-value (KV) cache mechanisms. However, they often suffer from the well-known error accumulation issue \cite{huang2025self,chen2024diffusion}, as each frame depends on potentially corrupted previous outputs, causing errors to propagate progressively. 

To mitigate error accumulation, recent works have adopted attention sink mechanisms that retain initial tokens in the KV cache \cite{liu2025rolling,shin2025motionstream,yang2025longlive}. Such a design largely recovers the performance of sliding window attention and alleviates long-horizon drifting. However, a new challenge arises: by consistently preserving initial tokens throughout generation, models develop a strong bias toward the starting frame, leading to over-attention on initial content. This manifests as diminished motion dynamics, where subsequent frames fail to evolve naturally, and frequent visual flashbacks that revert to the first frame's appearance. While classical distribution matching distillation \cite{yin2024one,yin2024improved} minimizes the divergence between student and teacher output distributions to transfer knowledge, this strategy struggles to address the over-attention issue. The degraded samples, despite their motion deficiencies, usually exhibit good visual quality and already fall close to the teacher distribution, making them difficult to distinguish and optimize.

In this paper, we propose \textbf{{\method}}, a novel framework with two key technical innovations to ensure both high visual and dynamic fidelity for efficient streaming video generation. During training, {\method} generates video chunks autoregressively by conditioning on previously self-generated outputs through KV cache mechanisms to bridge the train-test gap, following Self Forcing \cite{huang2025self}.
Instead of using static initial tokens as sink tokens in the KV cache, we introduce \textbf{EMA-Sink}, a novel state packaging mechanism for ultra-long video sequences. The core idea of EMA-Sink is to maintain fixed-size tokens initialized from initial frames while continuously updating them by fusing evicted tokens via exponential moving average as they exit the sliding window. Without additional cost, this design not only compresses effective global context to maintain attention performance, but also introduces recent dynamics to prevent over-attending to initial frames.
To better distill motion dynamics from teacher models, we introduce Rewarded Distribution Matching Distillation \textbf{(Re-DMD)}. Instead of treating all samples equally as in vanilla distribution matching distillation, Re-DMD is able to distinguish samples with diminished motion dynamics and prioritizes matching with samples exhibiting greater dynamics. To this end, Re-DMD uses a powerful vision-language model as reward function to rate samples according to their motion quality, then uses these scores to weight distribution matching gradients. This effectively biases the distribution matching toward high-quality regions while preserving high data fidelity, leading to enhanced motion dynamics in streaming video generation. Comprehensive experimental evaluation on both short and long video benchmarks demonstrates that {\method} achieves state-of-the-art video quality at 23.1 FPS on a single H100 GPU.

\section{Related Works}
\label{sec:relatedworks}
\paragraph{Autoregressive long video generation.}
Video diffusion models have advanced short video generation, yet most state-of-the-art models are limited to 5–10 second clips. To reduce the high cost of bidirectional denoising, recent studies have adopted autoregressive diffusion modeling for long video generation~\cite{henschel2025streamingt2v,song2025history,zhang2025packing,yuan2025lumos,gao2025longvie,li2025stable,lin2025autoregressive,yesiltepe2025infinity}. Among these, Pyramidal-flow employs multi-scale flow matching to alleviate computational burden~\cite{jin2024pyramidal}, while SkyReels-V2 integrates diffusion forcing~\cite{chen2024diffusion} with structural planning and multi-modal control~\cite{chen2025skyreels}. FAR combines short and long-term contexts via flexible positional encoding~\cite{gu2025long}, and MAGI-1 utilizes chunk-wise prediction for scalable autoregressive generation~\cite{teng2025magi}. CausVid reformulates bidirectional diffusion as causal generation through distribution matching distillation~\cite{yin2024one,yin2024improved} to reduce denoising steps~\cite{yin2025slow}. Self-Forcing builds on this framework to mitigate train-test discrepancy by simulating inference conditions~\cite{huang2025self}, which is further extended by LongLive through KV recaching and stream-based fine-tuning for long video generation~\cite{yang2025longlive}, and by Rolling-Forcing via joint denoising for simultaneous multi-frame processing~\cite{liu2025rolling}. However, these methods consistently trade off motion dynamics against visual quality, often introducing cumulative artifacts.

\vspace{0.2em}
\noindent\textbf{Reinforcement learning for video generation.}
Reinforcement learning~\cite{712192} addresses optimizing non-differentiable metrics and temporally extended outcomes, enabling video generation models to better align with human preferences. Research has diverged into two strands. The first develops specialized datasets and reward models~\cite{liu2025improving,qin2024worldsimbench,xu2024visionreward} for video evaluation. The second integrates RL algorithms into generation pipelines. Some approaches use rewards~\cite{yuan2024instructvideo,prabhudesai2024video} to directly supervise generative models, while direct preference optimization (DPO) methods~\cite{rafailov2023direct,lu2025inpo,lu2025smoothed} implicitly learn preferences from datasets without explicit reward modeling, showing strong robustness~\cite{liu2025videodpo,zhang2024onlinevpo}. Additionally, policy optimization~\cite{zhang2023adadiff} techniques, such as Self-Forcing++~\cite{cui2025self}, incorporate Flow-GRPO~\cite{liu2025flow} into DMD-distilled models to improve long-term temporal smoothness. However, this method depends on pre-distilled models, with performance inherently tied to base model.

\section{Method}
\label{sec:method}
\subsection{Preliminaries}
\noindent\textbf{Autoregressive video diffusion models.} In autoregressive video diffusion models, an $N$-frame sequence $\boldsymbol{x}^{1:N}$ follows $p(\boldsymbol{x}^{1:N})=\prod_{i=1}^{N}p(\boldsymbol{x}^{i}|\boldsymbol{x}^{<i})$. 
Self Forcing~\cite{huang2025self} introduces an autoregressive self-rollout mechanism aligning training with inference. During training, each frame $\boldsymbol{x}^{i}$ undergoes iterative denoising conditioned on previously generated clean frames and its noisy state, sampling from the autoregressive distribution. 
A few-step diffusion model $G_{\theta}$ approximates each conditional $p(\boldsymbol{x}^{i}|\boldsymbol{x}^{<i})$. Given timesteps $\{t_{0},t_{1},\cdots,t_{T}\}$, denoising at step $t_{j}$ for frame $i$ processes noisy frame $\boldsymbol{x}^{i}_{t_{j}}$ conditioned on $\boldsymbol{x}^{<i}$, then reintroduces controlled Gaussian noise via forward process $\Psi$ to yield $\boldsymbol{x}^{i}_{t_{j}}$ for the next step. The model distribution is: $p_{\theta}(\boldsymbol{x}^{i}|\boldsymbol{x}^{<i})=f_{\theta,t_{1}}\circ f_{\theta,t_{2}}\circ \cdots \circ f_{\theta,t_{T}}(\boldsymbol{x}^{i}_{t_{T}})$ where $f_{\theta,t_{j}}(\boldsymbol{x}^{i}_{t_{j}})=\Psi(G_{\theta}(\boldsymbol{x}^{i}_{t_{j}},t_{j},\boldsymbol{x}^{<i}),t_{j-1})$ and $\boldsymbol{x}^{i}_{t_{T}}\sim \mathcal{N}(0,\mathbf{I})$.
For longer sequences, LongLive~\cite{yang2025longlive} uses sink tokens~\cite{xiao2023efficient} with $p(\boldsymbol{x}^{i}|\boldsymbol{x}^{1},\boldsymbol{x}^{i-w+1:i-1})$ (window size $w$) to model $p(\boldsymbol{x}^{i}|\boldsymbol{x}^{<i})$, but over-relies on the initial frame, limiting dynamic variation and smooth transitions. While models can output multi-frame chunks per step~\cite{teng2025magi,yin2025slow,huang2025self}, we term each chunk a ``frame" for simplicity.

\vspace{0.5em}
\noindent\textbf{Distribution matching distillation.} DMD~\cite{yin2024improved,yin2024onestep} distills multi-step diffusion models into a few-step generator $G$ by minimizing reverse KL divergence between real $p_{\text{real}}(\boldsymbol{x})$ and generated distributions $p_{\text{fake}}(\boldsymbol{x})$ across timesteps:
\begin{equation}
\begin{split}
    \nabla_{\theta}\mathcal{L}_{\text{DMD}}&\triangleq \mathbb{E}_{t}(\nabla_{\theta} \mathbb{D}_{\text{KL}}(p_{\text{fake},t}(\boldsymbol{x}_{t})||p_{\text{real},t}(\boldsymbol{x}_{t})))\\
    & \approx -\mathbb{E}_{t}\Big(\int  (s_{\text{real}}(\Psi(G_{\theta}(\epsilon),t),t)\\
    &-s_{\text{fake}}(\Psi(G_{\theta}(\epsilon),t),t))\frac{\text{d}G_{\theta}(\epsilon)}{\text{d}\theta}\text{d}\epsilon\Big).
\end{split}
\label{eq:dmd}
\end{equation}
where $\epsilon \sim \mathcal{N}(0,\mathbf{I})$, $\Psi$ denotes forward diffusion at timestep $t$. In diffusion models, the score function is defined as:
\begin{equation}
\small
   s_{\text{real}}(\boldsymbol{x}_{t},t)=\nabla_{\boldsymbol{x}_{t}}\log p_{\text{real},t}(\boldsymbol{x}_{t})=-\frac{\boldsymbol{x}_{t}-\alpha_{t}\mu_{\text{real}}(\boldsymbol{x}_{t},t)}{\sigma_{t}^{2}},
\label{eq:score}
\end{equation}
where $\mu_{\text{real}}$ is the denoised estimate, and $\alpha_{t}, \sigma_{t}$ are noise schedule parameters~\cite{ho2020denoising,nichol2021improved,karras2022elucidating}. DMD freezes pre-trained $\mu_{\text{real}}$ (teacher) and updates $\mu_{\text{fake}}$ on generator outputs.

\vspace{0.5em}
\noindent\textbf{Reinforcement learning.} A unified RL~\cite{712192} fine-tuning objective is established by maximizing the evidence lower bound for optimal video generation $\boldsymbol{x}_{0}$, culminating in an RL objective that makes an explicit trade-off between reward maximization and fidelity to the original model:
\begin{equation}
    \mathcal{J}_{\text{RL}}(p,q) = \mathbb{E}\Big[\frac{r(\boldsymbol{x}_{0},\boldsymbol{c})}{\beta}-\log \frac{p(\boldsymbol{x}_{0}|\boldsymbol{c})}{q(\boldsymbol{x}_{0}|\boldsymbol{c})}\Big].
    \label{eq:rl}
\end{equation}
Here, $\boldsymbol{x}_{0}$ denotes the output, $r$ represents the reward model, $\boldsymbol{c}$ represents the conditioning input, $p$ and $q$ are distributions, and $\beta$ acts as the regularization term.

\begin{figure}[ht]
  \centering
   \includegraphics[width=1\linewidth]{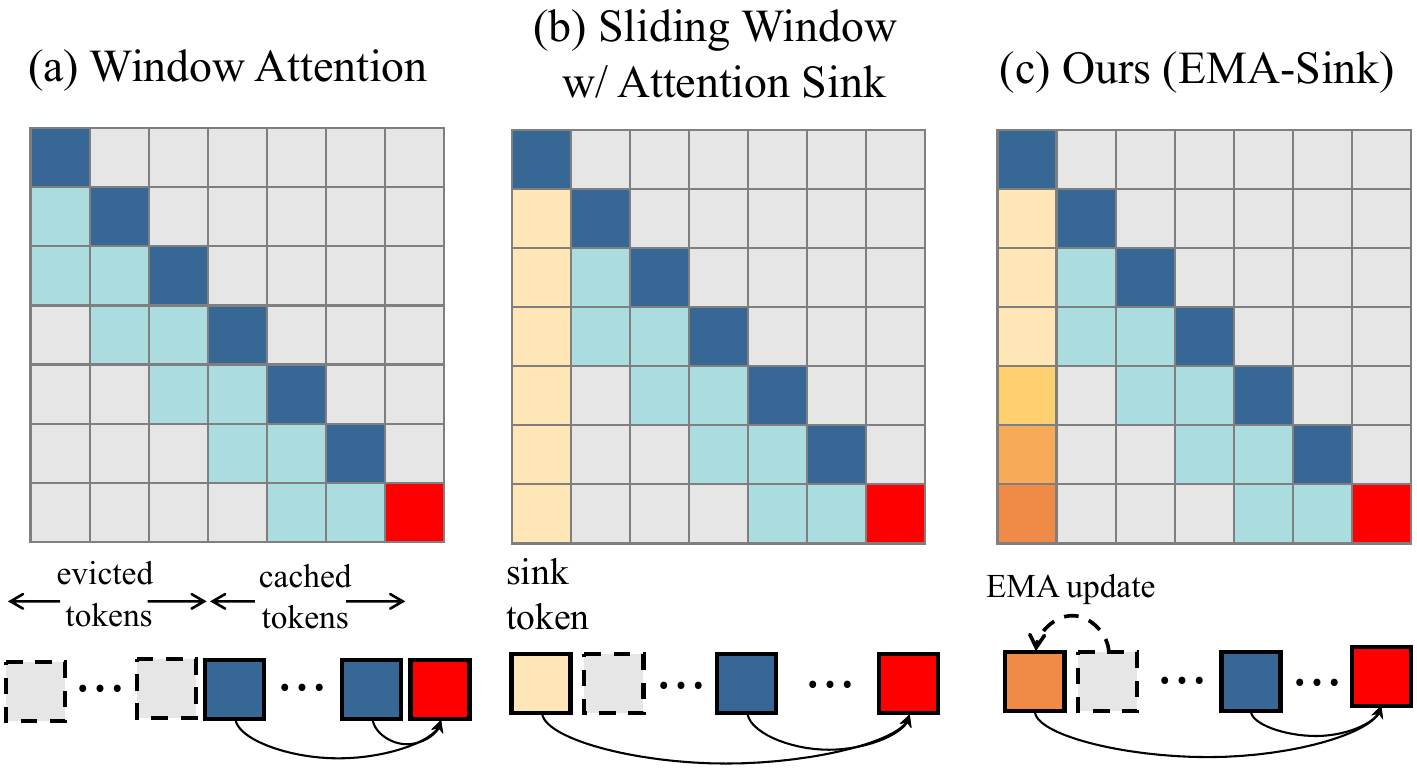}
   \caption{\textbf{Comparison of EMA Sink with Existing Methods.} Long video generation models typically extrapolate beyond their training sequence length during inference. (a) Window Attention caches only recent tokens for efficient inference but suffers performance degradation. (b) Sliding Window with attention sinks retains initial tokens for stable attention computation and recent tokens for extrapolation. However, discarding intermediate frames causes over-reliance on the first frame, leading to ``frame copying" and stiff transitions. (c) EMA Sink preserves full history through exponential moving average (EMA) updates of all historical frames, maintaining stable and consistent performance in long video extrapolation without increasing computational cost.}
   \label{fig:emasink}
\end{figure}

\begin{figure*}[ht]
  \centering
   \includegraphics[width=1\linewidth]{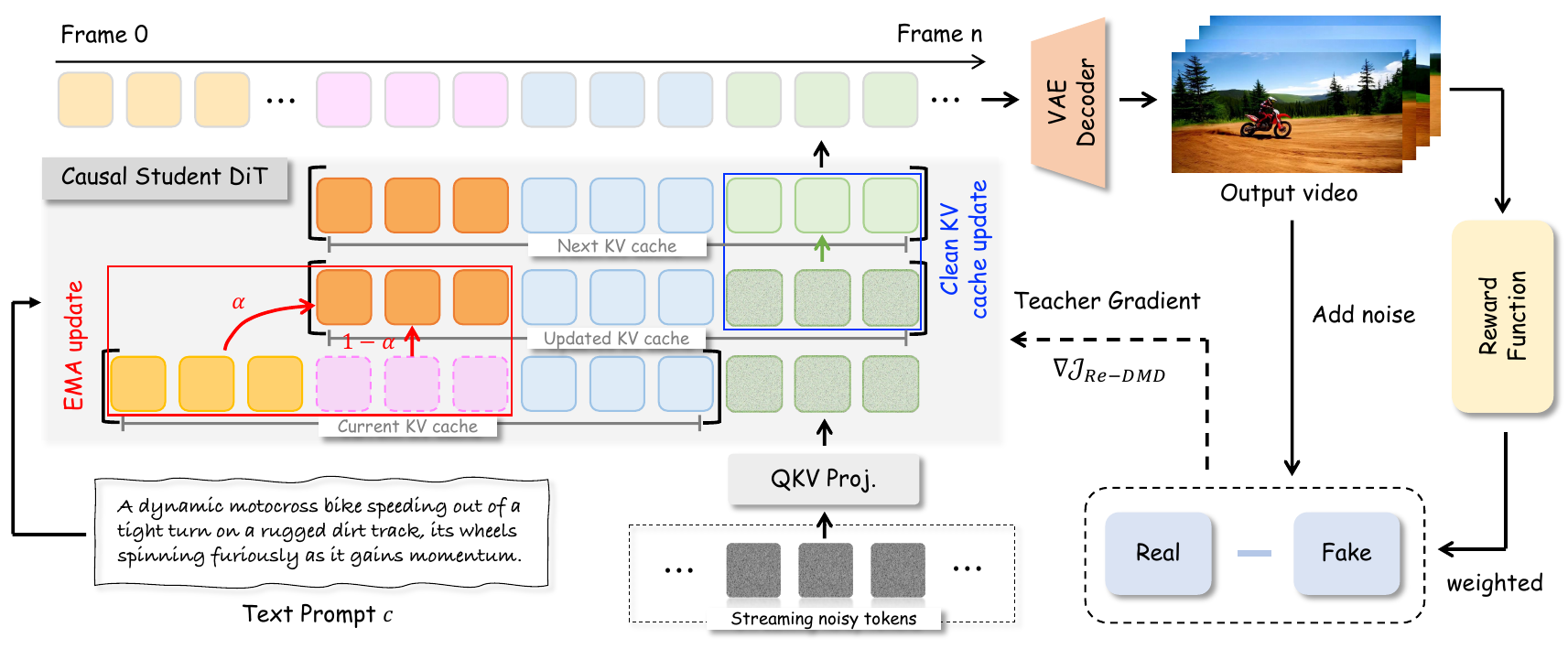}
   \caption{\textbf{Pipeline of {\method}.} In a streaming text-to-video generation, noisy tokens in the current stream are first projected to produce new key-value pairs (green blocks), which are appended to the KV cache for attention computation. When the current KV cache reaches its maximum attention window size, sink tokens initialized from start frames (yellow blocks) are updated via exponential moving average using evicted tokens (pink blocks). During training, hallucinated tokens are decoded into videos to compute a reward score via a reward function. This score is then used to weight the distribution matching gradient from the teacher model.}
   \label{fig:pipeline}
\end{figure*}

\subsection{EMA-Sink: state packaging for long video}
\vspace{0.5em}
\noindent\textbf{Problem formulation.} Efficient streaming video generation aims to create indefinitely long videos while maintaining strict temporal and causal consistency. Although sliding window attention is widely adopted in autoregressive models to reduce computational cost, current approaches fail to retain historical context beyond their limited attention windows~\cite{yin2025slow}. As generation progresses, earlier frames are discarded, creating an information bottleneck that diminishes global awareness and leads to temporal inconsistencies and quality drift over time. To address this, we introduce EMA-Sink, a novel state-packaging mechanism that compresses history to support efficient autoregressive generation. Our approach preserves global context in a compact, computationally efficient form throughout the streaming process.

For further illustration, given a noise schedule $\mathcal{T}=\{t_{j}\}_{j=0}^{T}$ consisting of distinct noise levels, the model processes each intermediate noisy frame $\boldsymbol{x}^{i}_{t_{j}}$ at denoising step $t_{j}$ and frame index $i$, incorporating earlier clean frames $\mathcal{X}^{i,w}=\big[\boldsymbol{x}^{i-w+1:i-1}\big]$ where $w$ denotes the window size used during video extrapolation ($i>w$). It first estimates a denoised version of the frame, then applies the forward diffusion operator $\Psi$ to reintroduce a lower level of Gaussian noise, producing $\boldsymbol{x}^{i}_{t_{j-1}}$ for subsequent denoising: $\Psi(G_{\theta}(\boldsymbol{x}^{i}_{t_{j}},t_{j},\mathcal{X}_{i}^{w}),t_{j-1})$, where $\boldsymbol{x}^{i}_{t_{T}}\sim \mathcal{N}(0,\text{I})$. 
As the window advances to frame $i+1$, the oldest frame $\boldsymbol{x}^{i-w+1}$ is removed from immediate access and is permanently discarded, thereby creating an information bottleneck~\cite{huang2025self}.

\vspace{0.5em}
\noindent\textbf{EMA-Sink mechanism.} Rather than discarding evicted frames, EMA-Sink maintains compressed global states $\boldsymbol{S}^{i}_{*}$ in the KV-cache through an exponential moving average. When frame $\boldsymbol{x}^{i-w}$ is evicted from the sliding window, its key-value pair ($\boldsymbol{K}^{i-w},\boldsymbol{V}^{i-w}$) is continuously fused into the compressed sink states $\boldsymbol{S}^{i}_{*}$: 
\begin{equation}
   \boldsymbol{S}^{i}_{\boldsymbol{K}} = \alpha \cdot \boldsymbol{S}^{i-1}_{\boldsymbol{K}} + (1-\alpha) \cdot \boldsymbol{K}^{i-w},
\end{equation}
\begin{equation}
   \boldsymbol{S}^{i}_{\boldsymbol{V}} = \alpha \cdot \boldsymbol{S}^{i-1}_{\boldsymbol{V}} + (1-\alpha) \cdot \boldsymbol{V}^{i-w}.
\end{equation}
Here $\alpha \in (0,1)$ is the momentum decay factor controlling compression rate, providing smooth temporal compression where recent information dominates while preserving a fading memory of distant history. During attention computation~\cite{vaswani2017attention}, we prepend the compressed sink states to the local window context: 
\begin{equation}   \boldsymbol{K}_{\text{global}}^{i}=\big[\boldsymbol{S}^{i}_{\boldsymbol{K}};\boldsymbol{K}^{i-w+1:i}\big],
\end{equation}
\begin{equation}
\boldsymbol{V}_{\text{global}}^{i}=\big[\boldsymbol{S}^{i}_{\boldsymbol{V}};\boldsymbol{V}^{i-w+1:i}\big],
\end{equation}
where $\boldsymbol{K}^{i-w+1:i}$ and $\boldsymbol{V}^{i-w+1:i}$ represent the key and value states from the current sliding window. This formulation allows each query to attend to both the fine-grained local context and the coarse-grained global history, effectively breaking the information bottleneck of the fixed window size. To handle the spatial-temporal nature of video while maintaining causal relationships, we employ a rotary position embedding (ROPE)~\cite{su2024roformer} when calculating attention. The position encoding is applied causally, ensuring that each position can only attend to previous positions in the sequence.

\begin{figure*}[ht]
  \centering
   \includegraphics[width=1\linewidth]{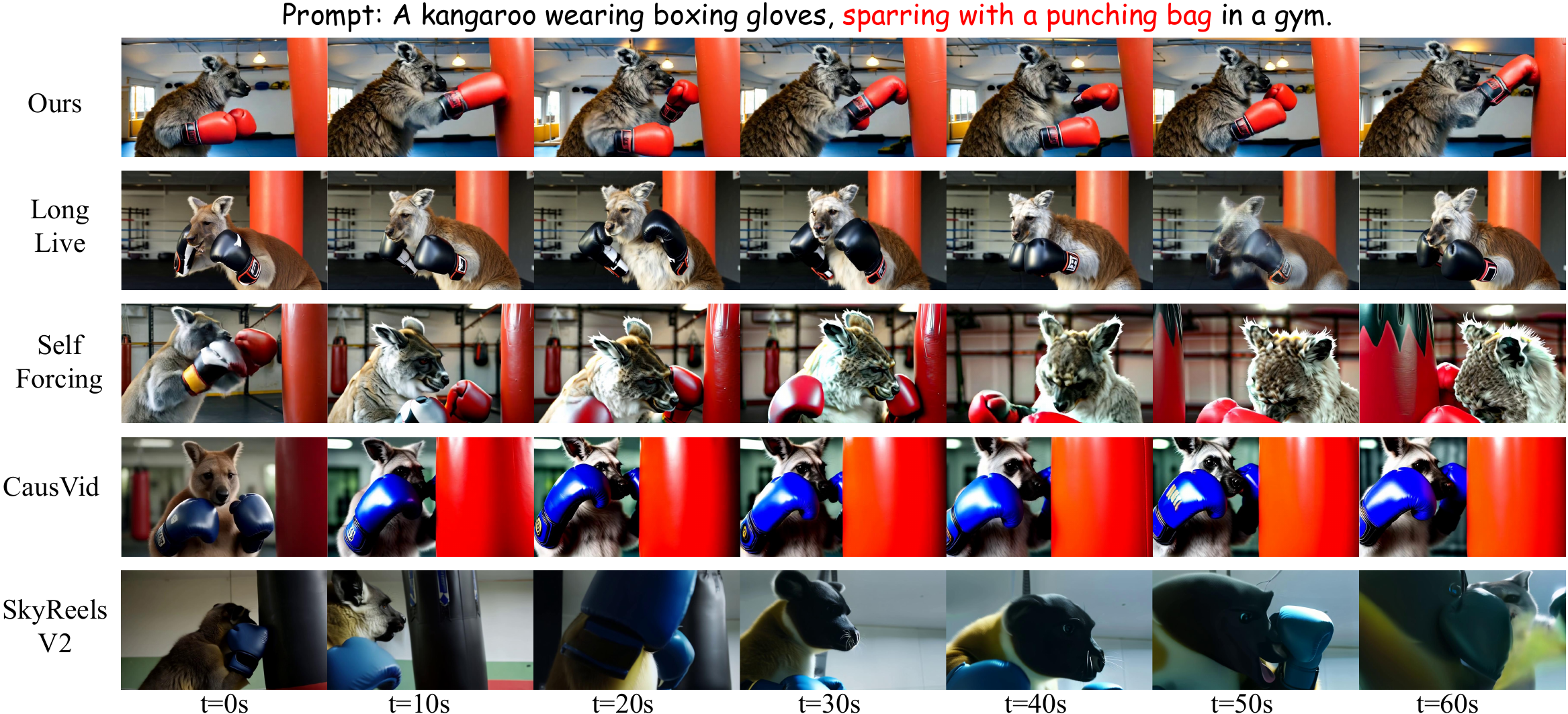}
   \caption{\textbf{Qualitative comparison on dynamic complexity of long video generation.} {\method} excels in both text alignment and motion dynamics while baselines exhibit diminished dynamics and weaker alignment.}
   \label{fig:qualitative1}
\end{figure*}

\subsection{Rewarded distribution matching distillation}
\paragraph{Problem formulation.} DMD~\cite{yin2024improved,yin2024onestep} offers an effective framework for converting multi-step diffusion models into efficient single-step generators by enforcing alignment between the fake and real distributions:
\begin{equation}
\begin{split}
    \mathcal{J}_{\text{DMD}}
    &=\mathbb{E}_{p(\boldsymbol{c})p_{\text{fake}}(\boldsymbol{x}_{0}|\boldsymbol{c})}\Big[\log\frac{p_{\text{fake}}(\boldsymbol{x}_{0}|\boldsymbol{c})}{p_{\text{real}}(\boldsymbol{x}_{0}|\boldsymbol{c})}\Big].
    \label{eq:logdmd}
\end{split}
\end{equation}
Despite its success in preserving sample fidelity, DMD has a fundamental limitation: it treats all regions of the target distribution uniformly, lacking any mechanism to prioritize high-quality outputs according to task-specific metrics. This becomes particularly problematic in video generation, where models progressively produce increasingly static frames during training. This observation motivates a key question: \textit{Can we incorporate motion awareness into the distillation process while maintaining distributional fidelity?} We address this challenge by integrating RL principles~\cite{712192} to bias the distillation toward high-reward regions of the output space, thereby generating content with enhanced properties without sacrificing data fidelity.

\vspace{0.5em}
\noindent\textbf{Re-DMD mechanism.}
We introduce Rewarded Distribution Matching Distillation (Re-DMD), which reweights the distribution matching objective according to sample motion quality. Our approach builds on the Reward-Weighted Regression framework~\cite{furuta2024improving,10.1145/1273496.1273590,lee2023aligning,liu2025improving}, which reformulates the reinforcement learning problem as probabilistic inference via the Expectation-Maximization (EM) algorithm~\cite{543975}. 

In the E-step~\cite{543975,10.1145/1273496.1273590}, we solve \cref{eq:rl} as a constrained optimization problem, obtaining the optimal solution:
\begin{equation}
    p(\boldsymbol{x}_{0}|\boldsymbol{c}) = \frac{1}{Z(\boldsymbol{c})}q(\boldsymbol{x}_{0}|\boldsymbol{c})\exp\Big(\frac{r(\boldsymbol{x}_{0},\boldsymbol{c})}{\beta}\Big),
\end{equation}
where $Z(\boldsymbol{c})=\sum_{\boldsymbol{x}_{0}}p(\boldsymbol{x}_{0}|\boldsymbol{c})\exp(\frac{r(\boldsymbol{x}_{0},\boldsymbol{c})}{\beta})$.We assign the distributions in \cref{eq:rl} as $q = p_{\text{fake}}'$ and $p = p_{\text{real}}'$. 

In the M-step~\cite{543975,10.1145/1273496.1273590}, we project the nonparametric optimal model $p = p_{\text{real}}'$ onto the parametric model by maximizing expected log-likelihood \cref{eq:rl} with respect to $p_{\text{fake}}$:
\begin{equation}
\small
\begin{split}
    \mathcal{J}_{\text{Re-DMD}} 
    &=\mathbb{E}_{p(\boldsymbol{c})p_{\text{fake}}'(\boldsymbol{x}_{0}|\boldsymbol{c})}\Big[\frac{\exp(r(\boldsymbol{x}_{0},\boldsymbol{c})/\beta)}{Z(\boldsymbol{c})}\log\frac{p_{\text{fake}}(\boldsymbol{x}_{0}|\boldsymbol{c})}{p_{\text{real}}(\boldsymbol{x}_{0}|\boldsymbol{c})}\Big].
\end{split}
\label{eq:rldmd}
\end{equation}
Computing the probability density to estimate this loss is generally intractable. However, when training the generator via gradient descent, we only need to obtain the gradient with respect to $\theta$. By differentiating \cref{eq:rldmd}, we obtain:

\begin{equation}
\small
\begin{split}
    \nabla_{\theta}\mathcal{J}_{\text{Re-DMD}} =&\mathbb{E}_{t}\Big(\nabla_{\theta}\mathbb{E}_{p_{\text{fake}}^{\boldsymbol{c}}(\boldsymbol{x}_{t})}\Big[\frac{\exp(r^{\boldsymbol{c}}(\boldsymbol{x}_{t})/\beta)}{Z(\boldsymbol{c})}\log\frac{p^{\boldsymbol{c}}_{\text{fake}}(\boldsymbol{x}_{t})}{p_{\text{real}}^{\boldsymbol{c}}(\boldsymbol{x}_{t})}\Big]\Big)\\
    \approx &-\mathbb{E}_{t}\Big(\int \exp(r^{\boldsymbol{c}}(\boldsymbol{x}_{t})/\beta)\cdot (s_{\text{real}}(\Psi(G_{\theta}(\epsilon),t),t)\\
    &-s_{\text{fake}}(\Psi(G_{\theta}(\epsilon),t),t))\frac{\text{d}G_{\theta}(\epsilon)}{\text{d}\theta}\text{d}\epsilon\Big).
\end{split}
\label{eq:gradientrldmd}
\end{equation}
where $\epsilon$ is random Gaussian noise, and $G_{\theta}$ is a generator parameterized by $\theta$. $s_{\text{real}}$ and $s_{\text{fake}}$ represent the score functions trained on the data and the generator's output distribution, respectively, using a denoising objective. In addition, $r^{\boldsymbol{c}}(\boldsymbol{x}_{t})$ is estimated by $r^{\boldsymbol{c}}(\boldsymbol{x}_{0})$. This approach stabilizes training and accelerates convergence by bypassing the intractable normalization constant and alleviating the need to compute the reward function's gradient.

\begin{figure*}[ht]
  \centering
   \includegraphics[width=1\linewidth]{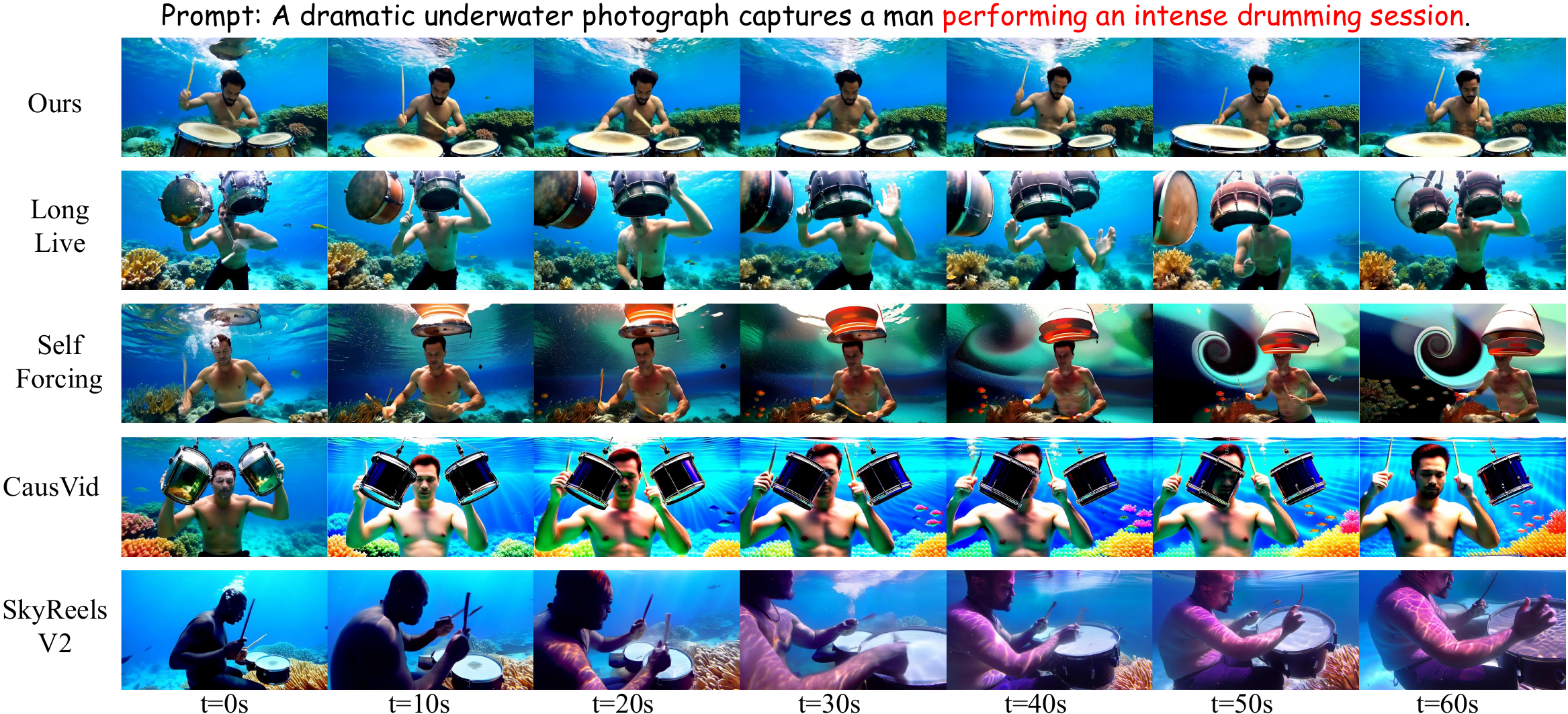}
   \caption{\textbf{Qualitative comparison on long-range temporal consistency.} {\method} maintains superior coherence over long-horizon, while baselines suffer from noticeable quality degradation and inconsistency over time.}
   \label{fig:qualitative2}
\end{figure*}

\subsection{Efficiency analysis}
\paragraph{Theoretical properties.}
The EMA-Sink enables token eviction in $O(1)$ time with low overhead. While attention remains $O(w^{2})$ in window size, it becomes independent of sequence length. By compressing history into a fixed-size sink, our method achieves constant memory usage relative to sequence length while retaining global context. The differentiable EMA enables gradient propagation through compression, supporting end-to-end learning of compression strategies. The Re-DMD objective implicitly optimizes a constrained reward maximization problem (maximizing expected reward under a distribution matching constraint), ensuring systematic quality improvement without distributional collapse. Notably, our approach avoids typical RL computational costs: the reward serves as a static weighting factor, eliminating backpropagation through reward models and preventing instability from noisy reward gradients.

\vspace{0.5em}
\noindent\textbf{Real-time long video inference.}
Long video generation faces quadratic complexity with dense causal attention, hindering real-time synthesis. Local window attention confines complexity to window size, independent of sequence length. With KV cache scaling by window dimension rather than video length, smaller windows accelerate inference and significantly improve efficiency.

\section{Experiments}
\label{sec:experiments}

\noindent\textbf{Implementation details.}
{\method} is built upon Wan2.1-T2V-1.3B~\cite{wan2025wan} to generate 5-second videos at $832 \times 480$ resolution. The model is first trained on 16k ODE solution pairs sampled from the base model, initialized with causal attention masking, following CausVid~\cite{yin2025slow}. Text prompts are drawn from the filtered and LLM-augmented VidProM~\cite{wang2024vidprom} dataset. We use VideoAlign's~\cite{liu2025improving} motion quality as the reward function with $\beta = \frac{1}{2}$. During training, denoising is applied chunk-wise using 3 latent frames per chunk, with denoising steps set to [1000, 750, 500, 250] and an attention window size of 9. Training runs for 600 steps on 64 H200 GPUs with a total batch size of 64 (\~ 3 hours). The AdamW optimizer is adopted with learning rates of $2.0 \times 10^{-6}$ for the generator $G_{\theta}$ and $4.0 \times 10^{-7}$ for the fake score $s_{\text{fake}}$, updating the generator every 5 steps and adjusting the fake score $s_{\text{fake}}$ accordingly.

\begin{table}[ht]
\small
\centering
\caption{\textbf{Short video performance comparison with baselines.} The comparison includes representative open-source models of comparable scale. Best results in \textbf{bold}, second-best \underline{underlined}.}
\begin{tabular}{lccccc}
\toprule
\multirow{2}{*}{Model} & \multirow{2}{*}{Params}& \multirow{2}{*}{FPS$\uparrow$} & \multicolumn{3}{c}{VBench evaluation scores $\uparrow$} \\
\cmidrule(lr){4-6} 
&&& Total & Quality & Semantic  \\
\midrule
\cellcolor{lightgray}\textit{Diffusion} &\cellcolor{lightgray}&\cellcolor{lightgray}&\cellcolor{lightgray}&\cellcolor{lightgray}&\cellcolor{lightgray} \\
LTX-Video~\cite{deng2024autoregressive} & 1.9B & 8.98 & 80.00 & 82.30 & 70.79 \\
Wan-2.1~\cite{wan2025wan} & 1.3B & 0.78 & 84.26 & 85.30 & 80.09 \\
\midrule
\cellcolor{lightgray}\textit{Autoregressive} &\cellcolor{lightgray}&\cellcolor{lightgray}&\cellcolor{lightgray}&\cellcolor{lightgray}&\cellcolor{lightgray} \\
SkyReels-V2~\cite{chen2025skyreels} &1.3B& 0.49 & 82.67 & \underline{84.70} & 74.53 \\
MAGI-1~\cite{teng2025magi} & 4.5B & 0.19 & 79.18 & 82.04 & 67.74 \\
NOVA~\cite{deng2024autoregressive} & 0.6B & 0.88 & 80.12 & 80.39 & 79.05 \\
Pyramid Flow~\cite{jin2024pyramidal} & 2B & 6.7 & 81.72 & 84.74 & 69.62\\
CausVid~\cite{yin2025slow} & 1.3B & 17.0 & 82.88 & 83.93 & 78.69 \\
Self Forcing~\cite{huang2025self} & 1.3B & 17.0 & \underline{83.80} & 84.59 & 80.64\\
LongLive~\cite{yang2025longlive}& 1.3B & 20.7 & 83.22 & 83.68 & \textbf{81.37} \\
Rolling Forcing~\cite{liu2025rolling} & 1.3B & 17.5 & 81.22 & 84.08 & 69.78\\
\midrule
Ours &1.3B& \textbf{23.1} & \textbf{84.13} & \textbf{84.84} & \underline{81.32} \\
\bottomrule
\end{tabular}
\label{tab:shortvideo}
\end{table}

\begin{table*}[h]
    \small
    \centering
    \caption{\textbf{Long video performance comparison with key baselines.} The best results are highlighted in \textbf{bold}.}
    \begin{tabular}{lccccccccccc}
        \toprule
        \multirow{2}{*}{Model}& \multicolumn{7}{c}{VBench Long Evaluation Scores $\uparrow$} & \multirow{2}{*}{Drift$\downarrow$} &  \multicolumn{3}{c}{Qwen3-VL Score $\uparrow$} \\
        \cmidrule(lr){2-8} \cmidrule(lr){10-12}
        &Total&Subject&Background&Smoothness&Dynamic&Aesthetic&Imaging Quality&&Visual&Dynamic&Text \\
        \midrule
        \cellcolor{lightgray}\textit{Diffusion Forcing}&\cellcolor{lightgray}&\cellcolor{lightgray}&\cellcolor{lightgray}&\cellcolor{lightgray}&\cellcolor{lightgray}&\cellcolor{lightgray}&\cellcolor{lightgray}&\cellcolor{lightgray}&\cellcolor{lightgray}&\cellcolor{lightgray}&\cellcolor{lightgray} \\
        SkyReels-V2~\cite{chen2025skyreels}&75.94&96.43&96.59&\textbf{98.91}&39.86&50.76&58.65&7.315&3.30&3.05& 2.70\\
        \midrule
        \cellcolor{lightgray}\textit{Distilled Causal}&\cellcolor{lightgray}&\cellcolor{lightgray}&\cellcolor{lightgray}&\cellcolor{lightgray}&\cellcolor{lightgray}&\cellcolor{lightgray}&\cellcolor{lightgray}&\cellcolor{lightgray}&\cellcolor{lightgray}&\cellcolor{lightgray}&\cellcolor{lightgray} \\
        CausVid~\cite{yin2025slow}&77.78&97.92&\textbf{96.62}&98.47&27.55&\textbf{58.39}&67.77&2.906&4.66&3.16& 3.32\\
        Self Forcing~\cite{huang2025self}&79.34&97.10&96.03&98.48&54.94&54.40&67.61&5.075&3.89&3.44& 3.11\\
        LongLive~\cite{yang2025longlive}&79.53&\textbf{97.96}&96.50&98.79&35.54&57.81&69.91&2.531&4.79&3.81&3.98\\
        \midrule 
        Ours&\textbf{81.41}&97.26&96.05&98.88&\textbf{66.95}&57.47&\textbf{70.06}&\textbf{2.505}& \textbf{4.82}& \textbf{4.18}& \textbf{4.04} \\
        \bottomrule
    \end{tabular}
    \label{tab:longvideo}
\end{table*}

\subsection{Comparison with state-of-the-art}
\paragraph{Short video generation.}
We generate 5-second videos using 946 official VBench~\cite{huang2023vbench,huang2024vbench++} prompts rewrited using Qwen/Qwen2.5-7B-Instruct~\cite{Qwen2.5-VL} following Self Forcing~\cite{huang2025self}, each sampled with 5 different seeds for comprehensive quality assessment. We benchmark our method against relevant open-source video generation models of comparable scale, including LTXVideo~\cite{deng2024autoregressive}, Wan2.1~\cite{wan2025wan}, SkyReels-V2~\cite{chen2025skyreels}, MAGI-1~\cite{teng2025magi}, CausVid~\cite{yin2025slow}, NOVA~\cite{deng2024autoregressive}, Pyramid Flow~\cite{jin2024pyramidal}, Self Forcing~\cite{huang2025self}, LongLive~\cite{yang2025longlive}, and Rolling Forcing~\cite{liu2025rolling}. The overall score of VBench comprises both quality and semantic components. As shown in \cref{tab:shortvideo}, our method achieves an overall score of 84.13 on the 5-second clips, surpassing all existing baselines and demonstrating superior video generation quality. Notably, our approach employs the smallest attention window while achieving the fastest inference speed among all compared methods. Specifically, we attain a real-time generation speed of 23.1 FPS, representing a $47.14\times$ speedup over SkyReels-V2 and a $1.36\times$ speedup over Self Forcing.

\vspace{0.5em}
\noindent\textbf{Long video generation.} 
Qualitative analysis of the results confirms the effectiveness of our approach. Specifically, \Cref{fig:qualitative1} illustrates its capability to produce more dynamic sequences for long video generation, and \Cref{fig:qualitative2} validates its improved temporal consistency. For long video quantitative evaluation, we use the first 128 prompts from MovieGen (consistent with CausVid~\cite{yin2025slow}), extending generation duration to 60 seconds. We employ VBenchLong~\cite{huang2024vbench++} metrics, including subject consistency, background consistency, motion smoothness, dynamic degree, aesthetic quality, and imaging quality, normalized and weighted using standard VBench~\cite{huang2023vbench} coefficients to compute the total score. To quantify drift in long video generation, we compute the standard deviation of imaging quality across 30 segments (2 seconds each) from 60-second videos. As shown in \cref{tab:longvideo}, our method achieves a total score of 81.41, significantly surpassing the state-of-the-art baseline LongLive (79.53). Notably, we observe substantial improvement in the dynamic metric (66.95), representing an 88.38\% boost in dynamic amplitude while minimizing quality drift, demonstrating our method's effectiveness with comparable performance on other metrics. Additionally, we employ Qwen3-VL-235B-A22B-Instruct~\cite{Qwen2.5-VL} to evaluate long video generation quality at 55–60 seconds, assessing visual quality, motion dynamics, and text alignment (see more details in the supplements). Each of the 128 videos is scored from 1 to 5, with averaged results showing our model achieves the best performance across all three metrics. We also include a user study for comprehensive comparison in the supplementary material, which demonstrates that our method consistently outperforms all key baselines.


\subsection{Ablation studies}
\begin{figure*}[ht]
  \centering
   \includegraphics[width=1\linewidth]{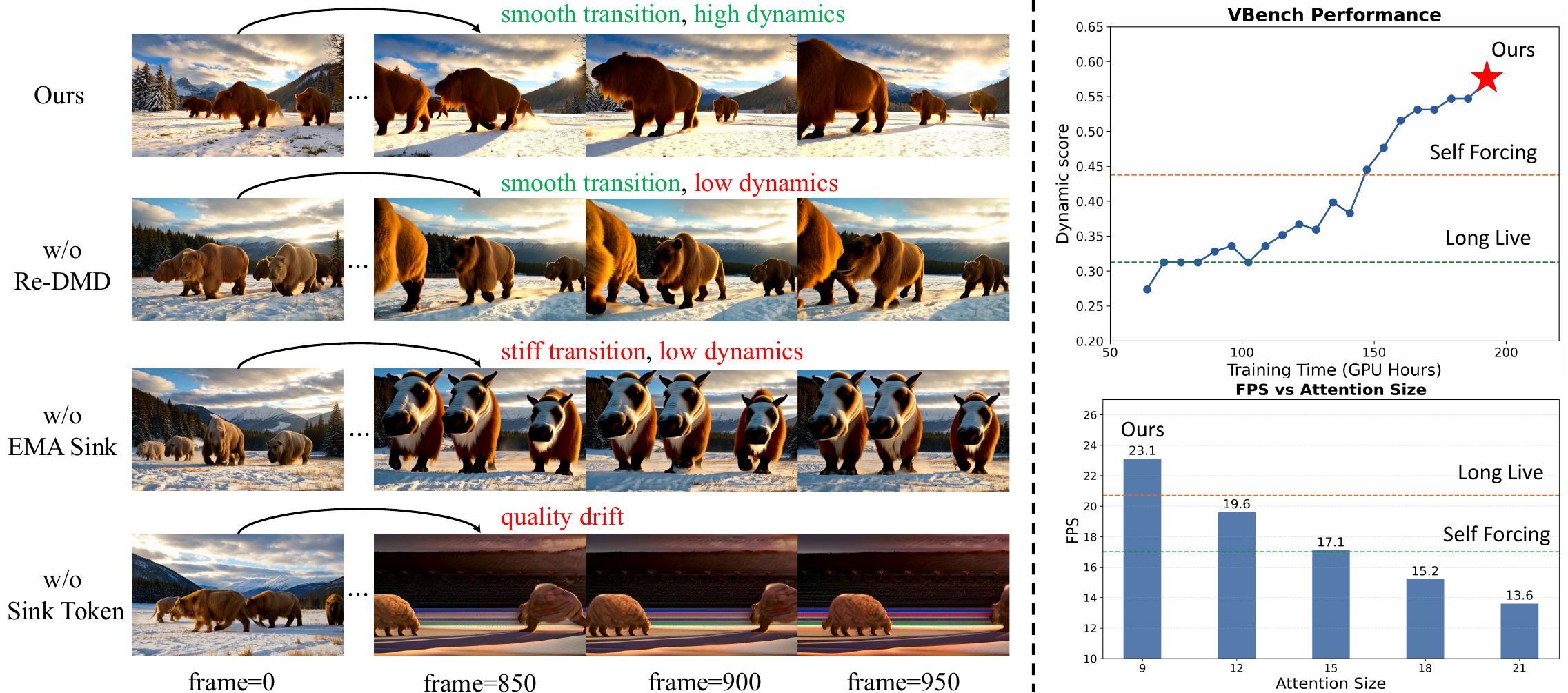}
   \caption{\textbf{(Left)} Ablation study on our proposed module, showing qualitative improvement. \textbf{(Right)} \textit{Top}: {\method} training leads to a steady rise in the dynamic score. \textit{Bottom}: The plot of attention size versus FPS underscores the source of our inference efficiency.}
   \label{fig:ablation}
\end{figure*}

\begin{table}[h]
\small
\centering
\caption{\textbf{Ablation studies on key components.} The best results for the ``Improvement" module are indicated in \textbf{bold}.}
\begin{tabular}{lccccc}
\toprule
\multirow{2}{*}{Model}& \multicolumn{4}{c}{VBench Evaluation Scores $\uparrow$} & \multirow{2}{*}{Drift$\downarrow$} \\
\cmidrule(lr){2-5} 
&Background & Smooth& Dynamic & Quality &  \\
\midrule
\cellcolor{lightgray}\textit{Improvement} &\cellcolor{lightgray}&\cellcolor{lightgray}&\cellcolor{lightgray}&\cellcolor{lightgray}&\cellcolor{lightgray} \\
Ours & 95.07 & 98.82 & \textbf{64.06} & 70.57 & 2.51 \\
w/o Re-DMD & \textbf{95.85} & \textbf{98.91} & 43.75 &\textbf{71.42}& \textbf{1.77} \\
w/o EMA & 95.61 & 98.64 & 35.15 & 70.50 & 2.65 \\
w/o Sink & 94.94 & 98.56 & 51.56 & 69.92 & 5.08 \\
\midrule
\cellcolor{lightgray}\textit{Impact of $\alpha$}&\cellcolor{lightgray}&\cellcolor{lightgray}&\cellcolor{lightgray}&\cellcolor{lightgray}&\cellcolor{lightgray} \\
$\alpha = 0.99$ & 95.90& 98.96& 65.15 & 70.81 & 2.52 \\
$\alpha = 0.9$ & 95.80 & 99.09 & 63.15 & 71.37 & 3.23 \\
$\alpha = 0.5$ & 94.57& 98.89& 64.37 &  71.11 & 3.78 \\
\midrule
\cellcolor{lightgray}\textit{Impact of $\beta$}&\cellcolor{lightgray}&\cellcolor{lightgray}&\cellcolor{lightgray}&\cellcolor{lightgray}&\cellcolor{lightgray} \\
$\beta = 1$ & 95.14& 98.31&  54.68& 71.73 &  2.63\\
$\beta = 2/3$ & 95.02& 98.46& 60.93 & 70.61 &  1.91\\
$\beta = 1/3$ & 94.94& 98.43&  58.59&  69.29&  2.02\\
$\beta = 1/5$ & 92.40& 96.40& 94.53 & 68.26 &  3.13\\
\bottomrule
\end{tabular}
\label{tab:ablation}
\end{table}

\paragraph{Impact of EMA-Sink and Re-DMD.} 
We show the effectiveness of {\method} through qualitative and quantitative comparisons. Qualitatively, as presented in \cref{fig:ablation}, our method maintains smooth transitions and high dynamism when generating 850–950 frames (approximately 1 minute), with clearly perceptible fluidity between consecutive frames. Without Re-DMD training, long video generation preserves high consistency with the initial frame and smooth scene transitions, but exhibits significantly reduced dynamism with the dynamic score drops from 64.06 to 43.75 ( \cref{tab:ablation}). As illustrated in \cref{fig:ablation}, removing the EMA Sink module results in considerable inconsistency with the first frame and minimal dynamism, reflected quantitatively by declining motion smoothness (98.91 to 98.64 in \cref{tab:ablation}) and dynamic score (43.75 to 35.15). Ablating the sink token leads to noticeable quality degradation.

\vspace{0.5em}
\noindent\textbf{Impact of EMA update weight $\alpha$.} 
An appropriately EMA coefficient $\alpha$ ensures smooth scene transitions in long videos, while a suitable $\alpha$ value effectively balances motion fluidity and temporal consistency. In our implementation, $\alpha$ is set to $9e^{-3}$. We can observe from \cref{tab:ablation} that $\alpha=0.99$ achieves a motion smoothness of 98.96 with a corresponding drift of 2.52. Conversely, reducing $\alpha$ to 0.9 improves motion smoothness to 99.09 but increases drift to 3.23.

\vspace{0.5em}
\noindent\textbf{Impact of reward weight $\beta$.}
The parameter $\beta$ modulates the reward term's influence, with smaller values assigning higher reward weight. As illustrated in \cref{tab:ablation}, an excessively small $\beta$ (e.g., 1/5) yields an overly high dynamic score (94.53) at the expense of background consistency (92.40), motion smoothness (96.40), and image quality (68.26). Conversely, an overly large $\beta$ (e.g., 1) produces an insufficient dynamic score (54.68). Therefore, we select $\beta=1/2$ to optimally balance these metrics.

\subsection{Analysis}
\paragraph{Dynamic enhancement of Re-DMD.}
We employ the VBench dynamic score (averaged over the first 128 prompts) to evaluate training effectiveness. An inspection of \cref{fig:ablation} reveals that the dynamic score increases steadily with training time while requiring modest computational resources (under 200 GPU hours). Our method surpasses LongLive (high consistency, low dynamism) after only 100 GPU hours and exceeds Self-Forcing (high dynamism, severe drift) after 150 GPU hours. Our model achieves high dynamism while maintaining strong consistency.

\vspace{0.5em}
\noindent\textbf{Impact of attention window size.}
The attention window size is a critical factor affecting the speed of real-time generation. \Cref{fig:ablation} demonstrates that inference FPS is inversely proportional to the size of the attention window.

\begin{figure}[t]
  \centering
   \includegraphics[width=1\linewidth]{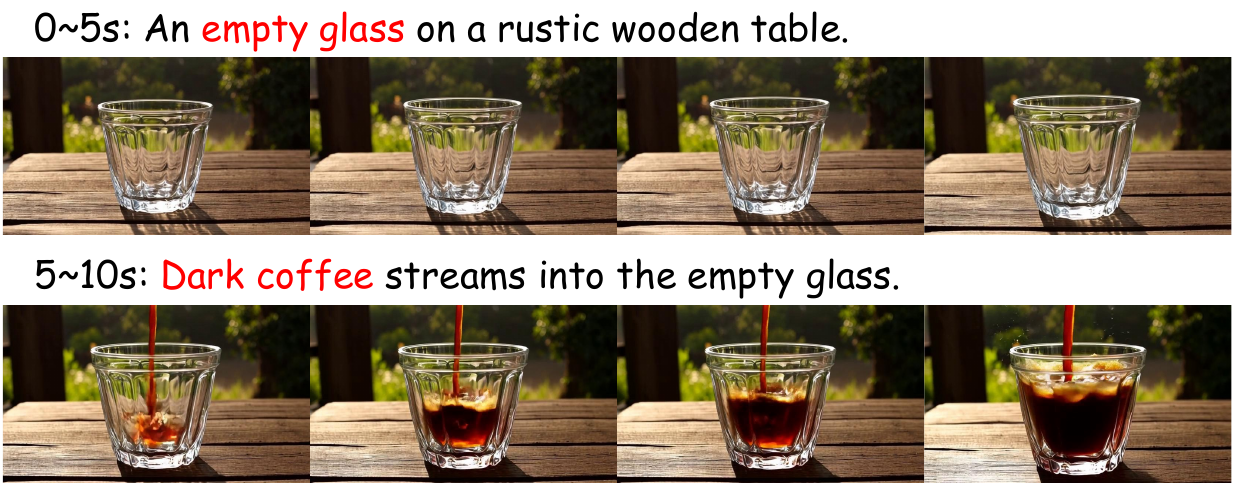}
   \caption{\textbf{Interactive video generation.} {\method} supports real-time prompt interaction with seamless transitions.}
   \label{fig:interactive}
\end{figure}

\vspace{0.5em}
\noindent\textbf{Interactive video generation.}
As shown in \cref{fig:interactive}, our method supports interactive video generation, allowing users to modify prompts during generation to control output content. This is achieved by clearing the previous cross-attention cache and recomputing it with the new prompt. Our EMA Sink mechanism ensures seamless prompt transitions while maintaining high temporal consistency.
\section{Conclusion}
\label{sec:conclusion}
We presented {\method}, which tackles the problem of motion stagnation in efficient streaming video generation. Our solution is built on two pillars: the EMA-Sink mechanism, which dynamically maintains context to prevent over-dependence on initial frames and ensures long-term coherence, and Re-DMD, which enhances motion dynamics by prioritizing high-reward samples during distillation. Our experiments confirmed that the proposed method achieves state-of-the-art performance on standard benchmarks. By successfully balancing high visual fidelity with strong dynamic motion, {\method} enables high-quality streaming video generation in real-time. This work establishes a new benchmark for performance and efficiency in generating dynamic, interactive virtual worlds.

\section{Acknowledgments}
This work was supported by the National Major Science and Technology Projects (the grant number 2022ZD0117000) and the National Natural Science Foundation of China (grant number 62202426). This work was supported by Ant Group Research Intern Program.

{
    \small
    \bibliographystyle{ieeenat_fullname}

}

\newpage
\clearpage
\setcounter{page}{1}

\setcounter{section}{0}

\onecolumn  
\begin{center}
    \textbf{\Large Reward Forcing: Efficient Streaming Video Generation with \\
    Rewarded Distribution Matching Distillation} \\[1ex]
    \Large Supplementary Material
\end{center}

\suppsection{More Video Results}

Please check the videos in the project page \url{https://reward-forcing.github.io/}. These videos are compressed to approximately 40\% of their original file size without significant quality degradation. 

\vspace{0.5em}
\noindent\textbf{Comparison with state-of-the-art methods.}
In addition to \cref{fig:teaser}, \cref{fig:qualitative1}, and \cref{fig:qualitative2} in the main paper, we provide additional video results in our supplementary materials for a more comprehensive evaluation of long videos (approximately 1 minute) generated by different methods. This page includes comparative studies with state-of-the-art methods. We randomly sample prompts from MovieGenBench~\cite{polyak2024movie}, focusing on Scene Navigation and Object Motion. As demonstrated in the videos, our {\method} preserves high visual fidelity while exhibiting superior motion dynamics over ultra-long horizon, which is crucial for simulating dynamic environments.

\vspace{0.5em}
\noindent\textbf{Interactive videos.}
In addition to \cref{fig:interactive} in the main paper, we include more interactive video results in the ``{\method}.html" page, demonstrating that our {\method} enables user interaction during streaming generation. Specifically, by switching prompts and resetting the cross-attention cache, the model can introduce new events into the ongoing video.

\suppsection{User Studies}
\label{sec:suppuserstudies}
\paragraph{Experimental setup.}
To comprehensively evaluate the performance of our proposed method in long video generation, we conducted a user study with 20 participants. Each participant was presented with 20 video groups, where each group contained four videos generated by different methods: CausVid~\cite{yin2025slow}, Self-Forcing~\cite{huang2025self}, LongLive~\cite{yang2025longlive}, and Reward Forcing (ours). The videos were randomly labeled as A, B, C, and D to avoid bias. In total, we collected 1,600 evaluations (20 participants $\times$ 20 video groups $\times$ 4 videos).

\paragraph{Evaluation protocol.}
Participants are asked to evaluate each video for three key criteria using a 4-point Likert scale (1-4):
\begin{itemize}
    \item Long-Range Temporal Consistency: This metric assesses whether the video maintains visual quality and coherence throughout its entire duration without experiencing visual drift, artifacts, or inconsistencies. Participants evaluated how well each video preserved semantic and structural consistency from start to finish.
    \item Dynamic Complexity: This metric measures the naturalness, richness, and engagement of motions and changes in the video. Participants assessed whether the generated content exhibited realistic and diverse dynamics rather than static or repetitive patterns.
    \item Overall Preference: This metric captures the holistic quality and appeal of each video, combining factors such as visual fidelity, coherence, motion quality, and subjective viewing experience.
\end{itemize}

For each criterion, participants assigned scores ranging from:

\begin{itemize}
    \item 4 (Good): High quality with no noticeable issues.
    \item 3 (Borderline Accept): Acceptable quality with minor issues.
    \item 2 (Borderline Reject): Below acceptable quality with noticeable issues
    \item 1 (Poor): Unacceptable quality with major issues.
\end{itemize}

\begin{figure}[h]
    \centering
    \includegraphics[width=1\linewidth]{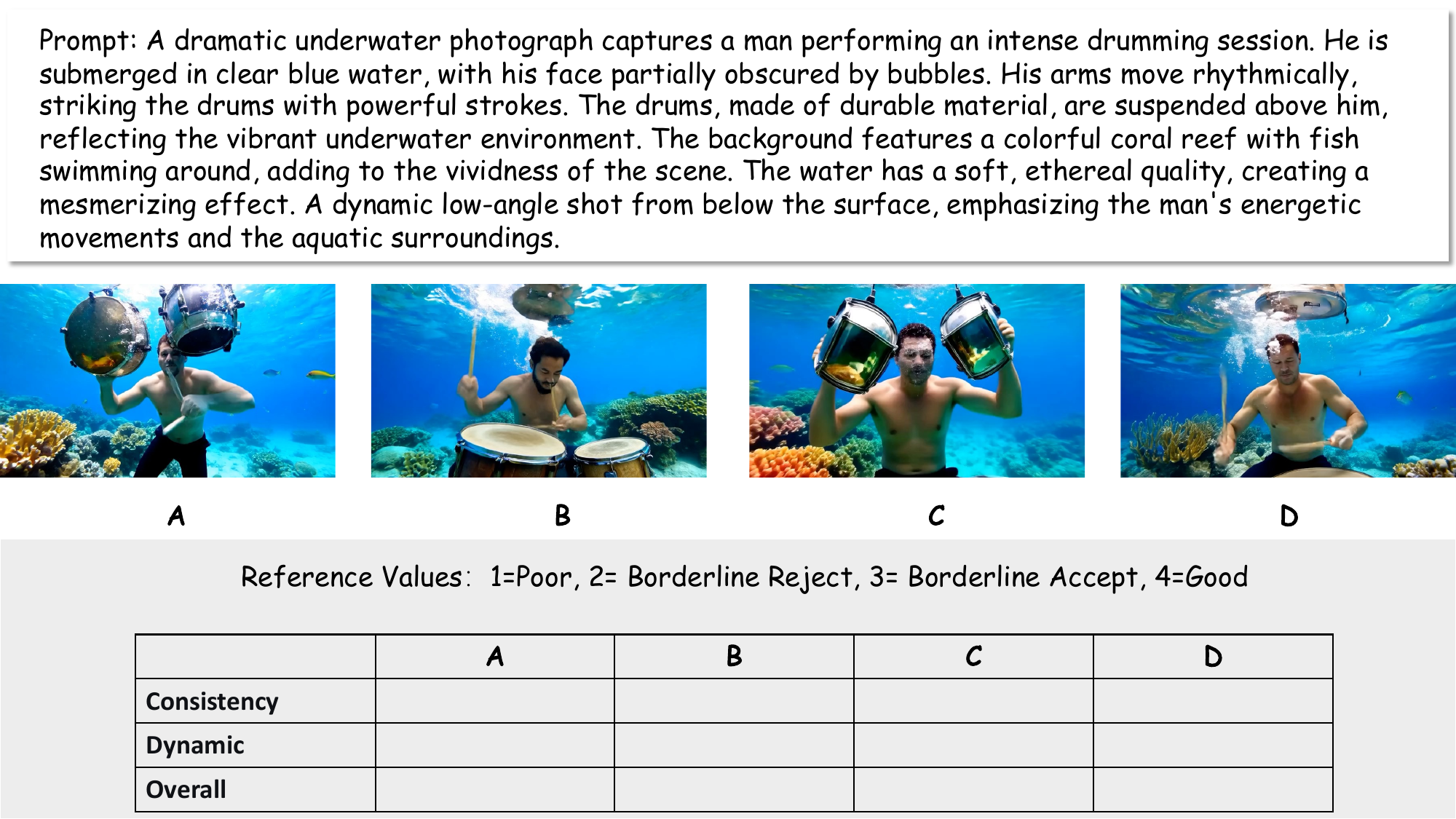}
    \caption{User study instruction screenshots.}
    \label{fig:suppscreenshots}
\end{figure}

\paragraph{Results and analysis.}
The user study results unequivocally demonstrate the superiority of our proposed Reward Forcing method over the baseline models across all evaluation criteria (\cref{tab:suppuserstudies}). Our method achieved the highest scores, nearing the ``Good" (4) benchmark on the Likert scale, with 3.60 for Temporal Consistency, 3.72 for Dynamic Complexity, and 3.75 for Overall Preference. This indicates that participants consistently rated our videos as high-quality with no noticeable issues. These results validate that Reward Forcing sets a new state-of-the-art for coherent and engaging long video generation.
\begin{table*}[h]
    \centering
    \caption{\textbf{Average User Rating.}}
    \begin{tabular}{lccc}
        \toprule
        Models & Temporal Consistency & Dynamic Complexity & Overall Preference\\
        \midrule
        CausVid~\cite{yin2025slow} &1.81408&1.72676& 1.87324\\
        Self Forcing~\cite{huang2025self} &1.19437&1.75493& 1.27042\\
        LongLive~\cite{yang2025longlive} &2.78873&2.38310&2.74648 \\
        \midrule
        Reward Forcing &\textbf{3.60282}&\textbf{3.72113}& \textbf{3.75493}\\
        \bottomrule
    \end{tabular}
    \label{tab:suppuserstudies}
\end{table*}

\suppsection{More Quantitative Results and Details}

\paragraph{Quality drift.} 
We report the quality drift evaluation results in \cref{tab:longvideo} and \cref{tab:ablation} in the main paper. To quantify the variability in long video imaging quality, we calculate the quality score drift along the temporal horizon using standard deviation, inspired by Zhang et al.~\cite{zhang2025packing}. Each one-minute video is divided into $M$ clips where $M=30$, each lasting 2 seconds. For any given long video clip $V_{i}$, we compute the drift as follows:
\begin{equation}
    \text{Drift}(V_{i}) = \sqrt{\frac{1}{M-1}\sum_{j=1}^{M}(s_{i,j}-\bar{s}_{i})},
    \label{suppeq:drift}
\end{equation}
where $C_{i,j}$ represent clip $j$ from video $i$, $s_{i,j}$ be the imaging quality score of clip $C_{i,j}$. The overall drift across all videos is the mean of individual video drifts:
\begin{equation}
    \text{Drift} = \frac{1}{N} \sum_{i=1}^{N}\text{Drift}(V_{i}),
    \label{suppeq:overalldrift}
\end{equation}
where $N$ is the total number of videos. Our results show that this metric effectively reflects video quality over long horizons, demonstrating a strong correlation between lower drift scores and more consistent visual fidelity throughout the sequences.

\vspace{0.5em}
\noindent\textbf{Qwen3-VL evaluation details.}
\label{sec:suppqwen3}
We use a powerful vision-language model, Qwen3-VL-235B-A22B-Instruct~\cite{yang2025qwen3}, for a more comprehensive evaluation and report the results in \cref{tab:longvideo} in the main paper. We include the evaluation template and detailed results for different methods as follows.

\begin{tcolorbox}[
  colback=black!5!white,
  title=Full Input Template,
]
\tboxsize
\textbf{\texttt{[VIDEO]}}
``````Please act as a video quality evaluation expert and rate the given video and text prompt on a scale of 1-5 across the following three dimensions:\\

**Evaluation Dimensions:**

1. **Text Alignment**:\\
Measures the consistency between the video content and the text description.\\
   - 1: Completely Irrelevant - Content is unrelated or severely contradicts the description.\\
   - 2: Mostly Mismatched - Only a few minor elements are relevant; the core concept is missing or incorrect.\\
   - 3: Partially Matched - The core idea is present but with significant deviations or missing key elements.\\
   - 4: Largely Consistent - Faithfully represents the description with only minor omissions or discrepancies.\\
   - 5: Perfectly Aligned - Comprehensive and accurate representation of the entire text description.\\

2. **Dynamics**: Evaluates the dynamism and fluidity of the entire scene, including camera movement, object motion, and scene transitions.\\
   - 1: Static / Disjointed - Little to no dynamic elements; or motion is severely broken and incoherent.\\
   - 2: Mostly Static - Limited, simple motion; dynamics feel stiff, mechanical, or poorly executed.\\
   - 3: Moderately Dynamic - Basic movement is present but lacks fluidity and natural flow; may appear robotic.\\
   - 4: Largely Dynamic - Generally fluid and engaging motion with a good sense of flow; minor imperfections may exist.\\
   - 5: Highly Dynamic - Exceptionally smooth, natural, and purposeful motion that enhances the visual narrative.\\

3. **Visual Quality**: Assesses the technical execution, including clarity, color grading, composition, and the absence of artifacts.\\
   - 1: Very Poor - Severely blurry, heavy visual artifacts (e.g., distortion, tearing), and/or extreme color issues (e.g., over-saturation, color banding).\\
   - 2: Poor - Consistently blurry, noticeable noise, unnatural color palette, or frequent minor artifacts.\\
   - 3: Fair - Passable clarity and color, but with visible technical flaws; composition may be unremarkable.\\
   - 4: Good - Clear and mostly sharp, with natural and balanced colors; good composition and only minor, infrequent issues.\\
   - 5: Excellent - Technically superior: sharp, well-composed, with vivid yet natural colors, and free from visible artifacts or distortions.\\

**Scoring Requirements:**\\
- Please output strictly in the following format, only numbers and brief reasons:\\
Text Alignment: [1-5]\\
Reason: [brief explanation]\\
Dynamics: [1-5]\\
Reason: [brief explanation]\\
Visual Quality: [1-5]\\
Reason: [brief explanation]\\

Now please evaluate the following content:\\
Text Prompt: "{}"\\
Video Content: Please carefully watch the provided video\\
"""

\end{tcolorbox}

\begin{tcolorbox}[
  colback=black!5!white,
  title=CausVid Full Results,
]
\tboxsize
Average Scores by Dimension:\\
Text Alignment: 3.32\\
Dynamics: 3.16\\
Visual Quality: 4.66\\

Detailed Scores by Dimension:\\
Text Alignment: [5, 1, 2, 4, 5, 5, 5, 5, 5, 1, 3, 4, 1, 1, 4, 2, 4, 4, 2, 2, 3, 3, 2, 1, 5, 3, 2, 5, 3, 4, 5, 4, 2, 3, 2, 1, 3, 3, 2, 3, 5, 3, 1, 4, 5, 2, 2, 3, 2, 4, 5, 5, 3, 5, 3, 5, 3, 5, 5, 5, 4, 2, 5, 5, 5, 5, 2, 5, 5, 5, 3, 2, 2, 5, 1, 3, 2, 4, 2, 2, 3, 2, 3, 3, 2, 1, 3, 2, 1, 2, 5, 3, 4, 4, 5, 3, 3, 5, 5, 1, 2, 4, 5, 5, 4, 3, 2, 5, 2, 5, 5, 3, 5, 4, 2, 5, 4, 2, 5, 5, 2, 1, 2, 5, 1, 3, 1, 5]\\
Dynamics: [4, 3, 3, 5, 4, 3, 4, 4, 3, 1, 4, 4, 2, 2, 4, 2, 4, 3, 4, 3, 4, 4, 3, 4, 3, 2, 3, 4, 2, 3, 4, 3, 3, 4, 2, 3, 4, 2, 2, 3, 4, 3, 3, 4, 3, 2, 3, 2, 4, 3, 3, 4, 4, 3, 4, 4, 3, 3, 4, 3, 4, 4, 4, 1, 3, 4, 1, 5, 3, 4, 4, 2, 3, 4, 3, 2, 3, 2, 4, 4, 5, 2, 3, 2, 2, 1, 4, 3, 4, 3, 3, 3, 2, 3, 4, 3, 2, 4, 4, 1, 1, 3, 4, 4, 4, 2, 4, 3, 3, 3, 3, 2, 4, 5, 3, 4, 4, 3, 3, 3, 3, 4, 2, 4, 2, 2, 3, 3]\\
Visual Quality: [5, 5, 5, 5, 5, 5, 5, 5, 5, 3, 5, 5, 5, 4, 5, 4, 5, 5, 4, 3, 5, 5, 4, 5, 5, 4, 4, 5, 5, 5, 5, 4, 5, 5, 5, 5, 5, 4, 5, 4, 5, 5, 5, 5, 4, 4, 4, 4, 5, 5, 5, 5, 5, 5, 5, 5, 4, 5, 5, 5, 5, 5, 5, 5, 4, 5, 4, 5, 5, 5, 5, 4, 4, 5, 4, 4, 4, 5, 5, 4, 4, 4, 4, 5, 5, 5, 5, 5, 5, 4, 5, 5, 5, 5, 5, 5, 5, 5, 5, 3, 3, 4, 5, 5, 5, 4, 4, 4, 4, 5, 5, 4, 5, 5, 5, 5, 5, 5, 5, 5, 5, 5, 3, 5, 5, 5, 4, 5]\\

\end{tcolorbox}

\begin{tcolorbox}[
  colback=black!5!white,
  title=SkyReels Full Results,
]
\tboxsize
Average Scores by Dimension:\\
Text Alignment: 2.70\\
Dynamics: 3.05\\
Visual Quality: 3.30\\

Detailed Scores by Dimension:\\
Text Alignment: [4, 2, 3, 2, 5, 5, 4, 2, 4, 1, 5, 3, 1, 2, 3, 1, 5, 2, 3, 1, 1, 1, 4, 1, 1, 2, 2, 1, 3, 4, 4, 1, 3, 3, 3, 1, 2, 5, 2, 2, 2, 2, 1, 4, 1, 1, 1, 2, 2, 2, 4, 3, 2, 5, 4, 5, 2, 5, 1, 5, 2, 2, 5, 2, 1, 5, 2, 4, 5, 3, 4, 2, 3, 3, 2, 4, 3, 1, 2, 2, 1, 3, 1, 3, 1, 5, 5, 1, 1, 3, 4, 4, 2, 1, 5, 4, 2, 3, 1, 4, 3, 1, 3, 2, 5, 3, 1, 1, 1, 5, 5, 1, 3, 3, 2, 3, 4, 1, 3, 5, 3, 2, 4, 1, 1, 5, 1, 5]\\
Dynamics: [5, 3, 1, 2, 3, 4, 3, 3, 3, 3, 4, 3, 3, 3, 4, 3, 5, 4, 4, 2, 3, 2, 4, 1, 3, 3, 4, 1, 2, 3, 3, 3, 2, 4, 4, 3, 3, 4, 3, 3, 2, 1, 3, 4, 4, 1, 2, 4, 3, 3, 5, 4, 2, 4, 3, 4, 3, 3, 2, 3, 3, 3, 4, 1, 2, 4, 3, 5, 3, 2, 4, 4, 4, 3, 3, 4, 3, 2, 2, 3, 1, 4, 3, 2, 2, 5, 4, 2, 1, 3, 3, 3, 2, 1, 3, 3, 3, 4, 3, 4, 3, 1, 4, 3, 5, 2, 1, 3, 1, 3, 4, 3, 4, 4, 2, 3, 4, 4, 2, 4, 4, 4, 5, 2, 4, 4, 3, 3]\\
Visual Quality: [4, 4, 2, 1, 4, 5, 3, 4, 3, 4, 5, 4, 4, 3, 4, 3, 5, 5, 3, 1, 3, 1, 4, 1, 3, 5, 4, 1, 4, 5, 4, 3, 5, 2, 4, 4, 3, 5, 4, 2, 3, 2, 3, 3, 4, 1, 3, 4, 3, 4, 4, 3, 2, 4, 3, 4, 3, 3, 3, 3, 2, 4, 3, 1, 3, 5, 3, 4, 3, 1, 4, 4, 4, 2, 3, 3, 4, 1, 3, 1, 3, 3, 2, 4, 4, 5, 3, 1, 2, 4, 4, 5, 3, 1, 4, 4, 4, 4, 2, 4, 3, 2, 4, 4, 5, 2, 3, 3, 1, 4, 4, 4, 3, 2, 2, 4, 3, 4, 2, 4, 3, 5, 4, 4, 5, 5, 4, 5]\\

\end{tcolorbox}
\begin{tcolorbox}[
  colback=black!5!white,
  title=Self Forcing Full Results,
]
\tboxsize
Average Scores by Dimension:\\
Text Alignment: 3.11\\
Dynamics: 3.44\\
Visual Quality: 3.89\\

Detailed Scores by Dimension:\\
Text Alignment: [4, 1, 2, 2, 5, 3, 5, 3, 5, 1, 4, 3, 1, 1, 5, 3, 2, 2, 3, 3, 1, 2, 2, 1, 5, 4, 2, 5, 4, 5, 3, 2, 3, 3, 2, 5, 1, 3, 3, 1, 4, 3, 1, 2, 1, 2, 1, 4, 4, 3, 1, 5, 5, 5, 1, 5, 4, 3, 3, 2, 4, 2, 5, 5, 2, 5, 3, 5, 5, 3, 4, 1, 3, 3, 2, 3, 3, 3, 3, 3, 3, 2, 2, 2, 5, 5, 4, 2, 1, 2, 5, 4, 2, 5, 5, 3, 2, 5, 1, 1, 3, 1, 5, 5, 3, 4, 2, 5, 2, 5, 4, 3, 3, 3, 3, 5, 2, 2, 5, 5, 3, 3, 4, 5, 3, 2, 1, 5]\\
Dynamics: [4, 3, 3, 2, 3, 3, 3, 3, 4, 3, 3, 3, 3, 3, 4, 3, 4, 4, 4, 4, 3, 4, 3, 1, 4, 3, 3, 4, 3, 4, 3, 4, 2, 4, 4, 5, 3, 3, 3, 3, 4, 2, 3, 2, 4, 4, 3, 4, 3, 3, 4, 4, 2, 3, 2, 4, 4, 3, 4, 3, 4, 4, 5, 3, 4, 4, 4, 5, 3, 3, 5, 3, 4, 4, 4, 4, 4, 2, 4, 4, 5, 3, 3, 2, 5, 5, 4, 4, 4, 4, 3, 3, 2, 4, 3, 3, 3, 4, 2, 2, 4, 1, 4, 4, 4, 3, 5, 3, 4, 3, 3, 4, 4, 4, 4, 3, 4, 3, 3, 3, 4, 4, 4, 4, 4, 3, 2, 3]\\
Visual Quality: [5, 4, 4, 2, 4, 4, 5, 4, 4, 4, 5, 3, 2, 3, 5, 4, 5, 4, 3, 4, 4, 3, 4, 2, 5, 4, 4, 4, 5, 5, 4, 4, 3, 3, 3, 5, 4, 4, 3, 2, 5, 4, 4, 3, 5, 3, 4, 4, 4, 3, 4, 5, 4, 4, 3, 3, 5, 4, 4, 3, 4, 5, 5, 4, 3, 4, 3, 5, 4, 4, 5, 4, 3, 4, 5, 5, 4, 1, 4, 3, 4, 2, 3, 4, 5, 5, 5, 4, 4, 5, 4, 5, 4, 4, 3, 3, 4, 4, 3, 4, 3, 1, 4, 4, 5, 5, 3, 3, 4, 3, 3, 4, 4, 5, 4, 5, 3, 3, 5, 4, 3, 5, 5, 4, 5, 4, 4, 4]\\

\end{tcolorbox}
\begin{tcolorbox}[
  colback=black!5!white,
  title=LongLive Full Results,
]
\tboxsize
Average Scores by Dimension:\\
Text Alignment: 3.98\\
Dynamics: 3.81\\
Visual Quality: 4.79\\

Detailed Scores by Dimension:\\
Text Alignment: [5, 2, 4, 5, 5, 5, 5, 5, 5, 1, 5, 3, 1, 2, 5, 4, 4, 4, 3, 5, 3, 3, 3, 4, 5, 5, 3, 5, 3, 5, 5, 4, 4, 3, 3, 2, 5, 5, 2, 3, 5, 3, 1, 5, 5, 5, 4, 5, 4, 1, 5, 5, 5, 5, 3, 5, 4, 5, 5, 5, 5, 2, 5, 5, 5, 3, 2, 5, 5, 5, 4, 1, 5, 5, 2, 5, 5, 5, 4, 4, 1, 4, 4, 2, 3, 5, 5, 3, 1, 5, 5, 3, 3, 5, 5, 4, 2, 3, 5, 3, 3, 5, 5, 5, 5, 5, 5, 5, 3, 5, 5, 5, 5, 5, 4, 5, 4, 3, 5, 5, 5, 2, 4, 5, 1, 3, 1, 5]\\
Dynamics: [5, 4, 3, 4, 4, 4, 3, 4, 3, 4, 4, 3, 4, 3, 4, 3, 5, 5, 4, 5, 3, 4, 4, 3, 4, 4, 4, 4, 2, 4, 4, 4, 4, 4, 4, 3, 4, 4, 3, 4, 4, 2, 3, 5, 4, 4, 5, 5, 4, 3, 4, 4, 1, 3, 4, 4, 4, 3, 4, 3, 5, 4, 4, 1, 4, 4, 4, 4, 3, 3, 4, 3, 5, 5, 4, 5, 4, 4, 5, 5, 4, 4, 3, 4, 2, 5, 4, 4, 4, 5, 3, 3, 2, 3, 3, 3, 2, 3, 4, 3, 4, 4, 4, 4, 5, 5, 5, 3, 4, 3, 4, 5, 5, 5, 5, 3, 5, 4, 4, 4, 5, 4, 5, 3, 4, 2, 4, 3]\\
Visual Quality: [5, 5, 5, 5, 5, 5, 5, 5, 5, 5, 5, 5, 5, 5, 5, 5, 5, 4, 4, 5, 4, 4, 5, 5, 5, 4, 5, 5, 5, 5, 5, 5, 5, 5, 4, 5, 5, 5, 4, 5, 5, 5, 5, 5, 5, 5, 4, 5, 5, 5, 4, 5, 5, 5, 5, 5, 5, 5, 5, 5, 5, 4, 5, 5, 5, 5, 4, 5, 5, 5, 5, 4, 5, 4, 3, 5, 5, 5, 5, 5, 4, 4, 4, 5, 4, 5, 5, 5, 5, 5, 5, 5, 4, 5, 5, 5, 5, 5, 5, 4, 5, 5, 4, 5, 5, 4, 4, 4, 5, 5, 5, 5, 5, 5, 5, 5, 5, 5, 5, 5, 4, 4, 5, 5, 5, 5, 5, 5]\\

\end{tcolorbox}
\begin{tcolorbox}[
  colback=black!5!white,
  title=Reward Forcing Full Results,
]
\tboxsize
Average Scores by Dimension:\\
Text Alignment: 4.04\\
Dynamics: 4.18\\
Visual Quality: 4.82\\

Detailed Scores by Dimension:\\
Text Alignment: [5, 2, 4, 2, 5, 5, 5, 5, 3, 1, 5, 5, 1, 1, 5, 5, 5, 4, 5, 5, 3, 5, 5, 4, 5, 5, 4, 5, 3, 5, 5, 4, 5, 4, 3, 2, 4, 4, 1, 3, 5, 3, 1, 5, 5, 5, 2, 5, 3, 2, 5, 5, 5, 5, 5, 5, 3, 5, 5, 5, 5, 2, 5, 5, 5, 2, 2, 5, 5, 5, 4, 1, 5, 5, 3, 5, 5, 5, 3, 4, 2, 4, 3, 4, 4, 5, 4, 5, 1, 5, 5, 4, 3, 5, 5, 4, 3, 5, 5, 2, 5, 4, 5, 5, 5, 4, 4, 5, 3, 5, 4, 5, 5, 2, 4, 5, 5, 3, 4, 5, 4, 2, 5, 5, 2, 5, 2, 5]\\
Dynamics: [5, 4, 3, 4, 4, 4, 4, 4, 4, 4, 4, 4, 4, 4, 4, 5, 5, 5, 4, 5, 3, 5, 4, 5, 4, 4, 4, 4, 4, 4, 4, 4, 4, 5, 4, 4, 4, 4, 3, 4, 4, 3, 4, 5, 5, 4, 4, 5, 4, 4, 4, 5, 4, 4, 4, 4, 4, 4, 4, 4, 5, 4, 4, 4, 4, 4, 3, 5, 4, 4, 5, 4, 5, 5, 4, 4, 5, 5, 4, 5, 4, 4, 4, 4, 3, 5, 4, 5, 4, 5, 4, 3, 3, 4, 4, 3, 4, 4, 4, 3, 5, 4, 4, 5, 5, 5, 5, 4, 4, 4, 3, 5, 5, 4, 5, 4, 5, 4, 4, 4, 5, 4, 5, 4, 4, 4, 4, 4]\\
Visual Quality: [5, 5, 5, 5, 5, 5, 5, 5, 5, 5, 5, 5, 5, 5, 5, 5, 5, 4, 4, 5, 4, 5, 5, 5, 5, 4, 5, 5, 5, 5, 5, 5, 5, 5, 4, 5, 5, 5, 4, 5, 5, 5, 5, 5, 5, 5, 4, 5, 5, 5, 4, 5, 5, 5, 5, 5, 5, 5, 5, 5, 5, 4, 5, 5, 5, 5, 4, 5, 5, 5, 5, 4, 5, 5, 3, 5, 5, 5, 5, 5, 4, 4, 4, 5, 4, 5, 5, 5, 5, 5, 5, 5, 4, 5, 5, 5, 5, 5, 5, 5, 5, 5, 4, 5, 5, 5, 4, 5, 5, 5, 5, 5, 5, 5, 5, 5, 5, 4, 5, 5, 4, 4, 5, 5, 5, 5, 5, 5]\\

\end{tcolorbox}

\paragraph{Detailed results on VBench \cite{huang2023vbench}.}
We report quantitative evaluation on VBench~\cite{huang2023vbench} using the extended prompts in \cref{tab:shortvideo} in the main paper. 
Specifically, the Quality Score is a weighted average of the following dimensions: subject consistency, background consistency, temporal flickering, motion smoothness, aesthetic quality, imaging quality, and dynamic degree. The Semantic Score is a weighted average of the following dimensions: object class, multiple objects, human action, color, spatial relationship, scene, appearance style, temporal style, and overall consistency. Each dimension's results are normalized using the following formula: normalized score = (score - min) / (max - min). The normalization range (minimum and maximum) for each dimension and the assigned weights used to compute the weighted average are provided in the \cref{tab:suppvbench}. 


\begin{table*}[ht]
\small
    \centering
    \caption{\textbf{Normalization ranges and weighting coefficients of VBench score.}}
    \begin{tabular}{lcccccccc}
        \toprule
        \multirow{2}{*}{}& Subject & Background & Temporal & Motion & Dynamic & Aesthetic & Imaging & Overall \\
        & Consistency & Consistency & Flickering & Smoothness & Degree & Quality & Quality & Consistency   \\
        \midrule
        \text{min}&0.1462&0.2615&0.6293&0.706&0.0&0.0&0.0&0.0 \\
        \text{max}&1.0&1.0&1.0&0.9975&1.0&1.0&1.0&0.364 \\
        \text{weighting coefficients}&1&1&1&1&0.5&1&1&1 \\
        \bottomrule
        \toprule
        & Object & Multiple & Human & \multirow{2}{*}{Color} & Spatial & \multirow{2}{*}{Scene}& Temporal & Appearance  \\
        & Class & Objects & Action && Relationship && Style & Style   \\
        \midrule
        \text{min}&0.0&0.0&0.0&0.0&0.0&0.0&0.0&0.0009 \\
        \text{max}&1.0&1.0&1.0&1.0&1.0&0.8222&0.364& 0.2855\\
        \text{weighting coefficients}&1&1&1&1&1&1&1&1 \\
        \bottomrule
    \end{tabular}
    \label{tab:suppvbench}
\end{table*}

In addition, we provide detailed evaluation in \cref{tab:suppshortvideoquality,tab:suppshortvideosemantic}. Our method achieves a total score of 84.13 on the short video generation task using the extended VBench prompts, consistently surpassing current state-of-the-art baselines and demonstrating its effectiveness.


\begin{table*}[h]
    \centering
    \caption{\textbf{Quality evaluation on extended VBench.}}
    \begin{tabular}{lccccccccc}
        \toprule
        \multirow{2}{*}{Model}& Subject & Background & Temporal & Motion & Dynamic & Aesthetic & Imaging & Quality & Total\\
        & Consistency & Consistency & Flickering & Smoothness & Degree & Quality & Quality & Score & Score \\
        \midrule
        CausVid~\cite{yin2025slow}&96.33&95.84&99.44&97.98&61.11&64.52&67.96&83.93&82.88 \\
        Self Forcing~\cite{huang2025self} &95.09&96.10&99.01&98.24&66.38&65.79&69.71&84.59& 83.80 \\
        LongLive~\cite{yang2025longlive} &96.98&96.92&99.35&98.79&40.83&67.03&69.18&83.68& 83.22\\
        \midrule
        Reward Forcing &95.43&96.59&98.97&98.32&68.05&65.66&69.38&\textbf{84.84}&\textbf{84.13} \\
        \bottomrule
    \end{tabular}
    \label{tab:suppshortvideoquality}
\end{table*}

\begin{table*}[h]
\small
    \centering
    \caption{\textbf{Semantic evaluation on extended VBench.}}
    \begin{tabular}{lccccccccccc}
        \toprule
        \multirow{2}{*}{Model}& Object & Multiple & Human & \multirow{2}{*}{Color} & Spatial & \multirow{2}{*}{Scene}& Temporal & Appearance & Overall & Semantic & Total\\
        & Class & Objects & Action && Relationship && Style & Style & Consistency & Score & Score \\
        \midrule
        CausVid~\cite{yin2025slow} &92.78&88.32&96.20&86.67&74.05&51.35&23.95&20.19&25.95&78.69& 82.88\\
        Self Forcing~\cite{huang2025self} &93.16&87.19&96.40&86.83&81.77&56.13&24.45&20.34&26.85&80.64& 83.80\\
        LongLive~\cite{yang2025longlive} &96.28&86.49&95.80&90.79&80.56&58.79&24.16&20.42&26.61&\textbf{81.37}& 83.22\\
        \midrule
        Reward Forcing &94.81&86.79&96.80&89.42&82.47&57.19&24.33&20.38&26.88&81.32&\textbf{84.13} \\
        \bottomrule
    \end{tabular}
    \label{tab:suppshortvideosemantic}
\end{table*}

\suppsection{More Implementation details}
\label{sec:suppdetails}
\paragraph{Noise schedule and model parameterization.}
Building upon the Wan2.1 and Self Forcing, our approach utilizes the flow matching framework. We implement a time step shift defined as $t'(k,t)=(kt/1000)/(1+(k-1)(t/1000))\cdot 1000$ with a shift factor $k$ set to 5. In the forward process, a sample is generated according to $x_{t}=\frac{t'}{1000}x+\frac{1-t'}{1000}\epsilon$, where $\epsilon$ is drawn from a standard normal distribution $\mathcal{N}(0,\mathbf{I})$ and $t$ ranges from 0 to 1000. The data prediction model is formulated as :
\begin{equation}
    G_{\theta}(\boldsymbol{x},t,c)=c_{\text{skip}}\cdot \epsilon - c_{\text{out}}\cdot v_{\theta}(c_{\text{in}}\cdot x_{t},c_{\text{noise}}(t'),c).
\end{equation}
The preconditioning coefficients remain consistent with the base models' settings: specifically, $c_{\text{skip}}$, $c_{\text{in}}$, $c_{\text{out}}$ are all $1$, and $c_{\text{noise}}(t)=t$. For our few-step diffusion sampling, we adopt a uniform 4-step schedule with time steps $\big[t_{1},t_{2},t_{3},t_{4}\big]=\big[1000,750,500,250\big]$.

\suppsection{Further Related Works}
\label{sec:supprelatedworks}
\paragraph{Video diffusion models.}
Video diffusion models~\cite{blattmann2023stable,he2022latent,ho2022imagen,ho2022video} have evolved from UNet~\cite{ronneberger2015u} backbones to Diffusion Transformers (DiTs)~\cite{peebles2023scalable}. Early approaches extended image diffusion models temporally~\cite{singer2022make} but lacked scalability. DiT's Transformer blocks enhance spatio-temporal modeling, enabling models like Sora~\cite{videoworldsimulators2024} and Hunyuan-Video~\cite{kong2024hunyuanvideo} to generate realistic, coherent videos. Hunyuan-Video integrates causal 3D VAE~\cite{kingma2013auto} and language models for textual control. Open-Sora~\cite{lin2024open} advanced efficiency and realism, while Wan 2.1~\cite{wan2025wan} validated large-scale pre-training benefits and CogVideoX~\cite{hong2022cogvideo,yang2024cogvideox} improved alignment via adaptive LayerNorm. For long video generation, Phenaki~\cite{villegas2022phenaki} uses discrete tokens, LVDM~\cite{he2022latent} employs hierarchical 3D latent generation, and NUWA-XL~\cite{yin2023nuwa} uses coarse-to-fine processing. LaVie~\cite{wang2025lavie} integrates rotary encoding and temporal attention, SEINE~\cite{chen2023seine} enables smooth transitions via stochastic masking, and LCT~\cite{guo2025long} extends to multi-shot generation. Diffusion forcing~\cite{chen2024diffusion} combines diffusion quality with autoregressive efficiency. StreamingT2V~\cite{henschel2025streamingt2v} adds memory modules, History-guided video~\cite{song2025history} uses historical context, FramePack~\cite{zhang2025packing} compresses frames, Lumos-1~\cite{yuan2025lumos} integrates LLM-like architecture, and LongVie~\cite{gao2025longvie} introduces multi-modal guidance and degradation-aware training. Test-time training methods~\cite{dalal2025one} generate minute-long videos but incur high costs. Training-free methods—RIFLEx~\cite{zhao2025riflex} adjusting positional embeddings, FreeNoise~\cite{qiu2023freenoise} combining noise rescheduling with windowed attention, and FreeLong~\cite{lu2024freelong} integrating multi-frequency information. 

\paragraph{Reinforcement learning for video models.}
Video generative models~\cite{chen2025longanimation,kim2025vip,sun2025swiftvideo,wu2025rewarddance,team2025longcat,vaswani2017attention,radford2021learning,cao2025dimension} using MLE or reconstruction loss misalign with human preferences. RL enables direct optimization of preference-aligned objectives~\cite{lin2025contentv,gao2025seedance}. Direct Preference Optimization (DPO)~\cite{furuta2024improving} dominates post-training alignment, including VideoDPO~\cite{liu2025videodpo} for temporal consistency, VisionReward~\cite{xu2024visionreward} for multi-objective preferences, and variants with physics-based generation. Group Relative Policy Optimization (GRPO), extending PPO~\cite{zhao2024dartcontrol,ma2024video,hu2023reinforcement}, improves generalization as shown in DanceGRPO~\cite{xue2025dancegrpo}. Reward-based approaches like InstructVideo~\cite{yuan2024instructvideo} with pretrained reward feedback and VADER~\cite{prabhudesai2024video} with unified differentiable rewards bypass policy learning. Inference-time methods like InfLVG~\cite{fang2025inflvg} incorporate GRPO for dynamic long-form optimization. Collectively, RL serves as both post-training alignment and structural component for preference-aware generation~\cite{feng2024matrix,gao2024flip,he2025pre,wang2025unified,cheng2025vpo}, bridging surrogate objectives and human-valued quality.

\suppsection{Discussion and Future Work}
\label{sec:suppdiscussion}
\paragraph{Generalizability.} 
Our method is designed to be general-purpose and plug-and-play, enabling seamless integration with various video generation architectures without requiring substantial modifications to existing pipelines. This flexibility represents a significant practical advantage, as it allows researchers and practitioners to adopt our approach with minimal overhead. 

\paragraph{Misalignment between reward functions and evaluation criteria.}
The first factor contributing to inconsistent performance is the misalignment between our reward function's optimization direction and VBench's evaluation criteria. VBench employs comprehensive metrics including temporal consistency, motion smoothness, subject consistency, background quality, aesthetics, and semantic alignment. Our reward model may prioritize certain dimensions over others—for example, heavily weighting temporal coherence while underemphasizing aesthetic qualities. This asymmetric optimization creates scenarios where reward improvements don't translate proportionally to VBench score gains. 

\paragraph{Video reward models.}
Our experiments show that current reward models can effectively guide quality improvements, as reflected in our competitive performance across multiple benchmarks. However, video reward models still face challenges in capturing certain nuanced aspects of video quality, such as long-range temporal dependencies, subtle temporal artifacts like frame jitter, and complex semantic attributes. These models are typically trained on datasets with subjective annotations that may not fully represent all quality dimensions. As video reward models continue to advance, our framework will naturally benefit from these improvements, enabling further optimization.

\paragraph{Future research directions.}
Future research should develop more sophisticated reward models capturing video quality nuances. Promising directions include: multi-objective reward modeling with separate components for different quality dimensions; hierarchical models assessing quality at multiple temporal scales; human-in-the-loop feedback mechanisms grounding models in perceptual judgments; domain-adaptive models adjusting criteria by content type; and architectures encoding physical and semantic priors about real-world dynamics. Advancing reward modeling along these dimensions could help our method achieve its full potential and demonstrate substantial, consistent improvements across comprehensive evaluation frameworks.

\suppsection{Border Social Impact}
\label{sec:suppimpact}
This work on efficient streaming video generation presents both significant opportunities and risks. On the positive side, the reduced computational requirements could democratize access to video synthesis technology, benefiting educational content creators, small organizations, and researchers with limited resources. The improved efficiency also reduces energy consumption, contributing to more sustainable AI development. However, we acknowledge serious concerns regarding potential misuse. The accessibility and speed of our method lowers barriers for creating deepfakes and misleading visual content that could spread misinformation or enable identity fraud. Additionally, our reward-based prioritization of dynamic content may inadvertently amplify biases present in vision-language models, potentially marginalizing underrepresented groups or activities. Questions of copyright infringement and consent regarding training data and generated likenesses also warrant careful consideration. To mitigate these risks, we strongly advocate for implementing digital watermarking and provenance tracking in any deployment of this technology. We encourage development of detection tools for synthetic content, clear content labeling practices, and robust usage policies prohibiting malicious applications such as non-consensual deepfakes. We support collaborative efforts among researchers, policymakers, and civil society to establish ethical guidelines, legal frameworks, and media literacy initiatives. Effective governance requires not only technological safeguards but also transparent data practices, diverse evaluation metrics to reduce bias, and ongoing dialogue about responsible use of generative video technologies.

\end{document}